\documentclass[journal,10pt,twocolumn,twoside]{IEEEtran}
\bstctlcite{IEEEexample:BSTcontrol}
\usepackage{setspace}
\usepackage{times,amsmath,amssymb}
\usepackage{slashbox}
\usepackage{graphicx}
\usepackage{xcolor}
\usepackage{setspace}
\usepackage{acronym}
\usepackage{algpseudocode}
\usepackage{algorithm}
\usepackage{booktabs}

\usepackage{multirow}

\newcommand{\algrule}[1][.2pt]{\par\vskip.5\baselineskip\hrule height #1\par\vskip.5\baselineskip}
\newcommand{\algcomment}[1]{\Statex$\%$ \emph{#1}}

\newcommand{\datax}{\mathbf{x}}
\newcommand{\dataxi}{\mathbf{x}_i}
\newcommand{\data}{\mathbf{\phi}}
\newcommand{\datai}{\mathbf{\phi}_i}
\newcommand{\datakj}{\mathbf{\phi}_{kj}}
\newcommand{\datacj}{\mathbf{\phi}_{cj}}
\newcommand{\centroidk}{\mathbf{\bar{\phi}}_{k}}
\newcommand{\centroidc}{\mathbf{\bar{\phi}}_{c}}
\newcommand{\centroidp}{\mathbf{\bar{\phi}}_{p}} 
\newcommand{\centroidn}{\mathbf{\bar{\phi}}_{n}} 
\newcommand{\Data}{\mathbf{\Phi}}

\newcommand{\tildeData}{\tilde{\mathbf{\Phi}}}
\newcommand{\pdata}{\mathbf{z}} 
 
\newcommand{\noDim}{D} 
\newcommand{\noNPTDim}{r} 
\newcommand{\nodim}{d} 
\newcommand{\noItems}{N}
\newcommand{\noClust}{K}
\newcommand{\Sx}{\mathbf{S}_x}
\newcommand{\St}{\mathbf{S}_t}
\newcommand{\Sp}{\mathbf{S}_p}
\newcommand{\Sn}{\mathbf{S}_n}
\newcommand{\tildeSx}{\tilde{\mathbf{S}}_x}
\newcommand{\tildeSt}{\tilde{\mathbf{S}}_t}
\newcommand{\tildeSp}{\tilde{\mathbf{S}}_p}
\newcommand{\tildeSn}{\tilde{\mathbf{S}}_n}
\newcommand{\Sb}{\mathbf{S}_b}
\newcommand{\Sw}{\mathbf{S}_w}
\newcommand{\Snw}{\mathbf{S}_{nw}}
\newcommand{\Snb}{\mathbf{S}_{nb}}
\newcommand{\SVDU}{\mathbf{U}}
\newcommand{\SVDS}{\mathbf{\Sigma}}
\newcommand{\SVDV}{\mathbf{V}}
\newcommand{\SVDUU}{\mathbf{P}}
\newcommand{\SVDSS}{\mathbf{\tilde{\Sigma}}}
\newcommand{\SVDVV}{\mathbf{Q}}
\newcommand{\Null}{\mathcal{N}}
\newcommand{\Nullt}{\mathcal{N}_t}
\newcommand{\Nullp}{\mathcal{N}_p}
\newcommand{\Nulln}{\mathcal{N}_n}
\newcommand{\Nullnb}{\mathcal{N}_{nb}}
\newcommand{\Nullnw}{\mathcal{N}_{nw}}
\newcommand{\Nullpm}{\mathbf{N}_p}
\newcommand{\Nullnm}{\mathbf{N}_n}
\newcommand{\Proj}{\mathbf{G}}
\newcommand{\proj}{\mathbf{g}}
\newcommand{\Projj}{\mathbf{W}}
\newcommand{\Projjj}{\mathbf{R}}
\newcommand{\projj}{\mathbf{w}}
\newcommand{\Mapp}{\mathbf{M}}
\newcommand{\mapp}{\mathbf{m}}
\newcommand{\K}{\mathbf{K}}

\newcommand{\I}{\mathbf{I}} 
\newcommand{\R}{\mathbb{R}}

\newcommand{\rank}{\mathrm{rank}}
\newcommand{\nullity}{\mathrm{nullity}}
\newcommand{\trace}{\mathrm{tr}}

\newcommand\blfootnote[1]{%
  \begingroup
  \renewcommand\thefootnote{}\footnote{#1}%
  \addtocounter{footnote}{-1}%
  \endgroup
}

\title{Null Space Analysis for\\Class-Specific Discriminant Learning}
\author{
\IEEEauthorblockN{Jenni Raitoharju\IEEEauthorrefmark{1}, \textit{Member, IEEE}, Alexandros Iosifidis\IEEEauthorrefmark{2}, \textit{Senior Member, IEEE}\\} 
\IEEEauthorblockA{\IEEEauthorrefmark{1} Unit of Computing Sciences, Tampere University, Finland, email: jenni.raitoharju@tuni.fi\\} 
\IEEEauthorblockA{\IEEEauthorrefmark{2} Department of Engineering, ECE, Aarhus University, Denmark, email: alexandros.iosifidis@eng.au.dk}} 
				
\normalsize
\begin{document}

\newacro{PCA}[PCA]{Principal Component Analysis}
\newacro{LDA}[LDA]{Linear Discriminant Analysis}
\newacro{KDA}[KDA]{Kernel Discriminant Analysis}
\newacro{CSDA}[CSDA]{Class-Specific Discriminant Analysis}
\newacro{NLDA}[NLDA]{Null space Linear Discriminant Analysis}
\newacro{ULDA}[ULDA]{Uncorrelated Linear Discriminant Analysis}
\newacro{OLDA}[OLDA]{Orthogonal Linear Discriminant Analysis}
\newacro{ROLDA}[ROLDA]{Regularized Orthogonal Linear Discriminant Analysis}
\newacro{NCSDA}[NCSDA]{Null space Class-Specific Discriminant Analysis}
\newacro{OCSDA}[OCSDA]{Orthogonal Class-Specific Discriminant Analysis}
\newacro{UCSDA}[UCSDA]{Uncorrelated Class-Specific Discriminant Analysis}
\newacro{HNCSDA}[HNCSDA]{Heterogeneous Null space Class-Specific Discriminant Analysis}
\newacro{HOCSDA}[HOCSDA]{Heterogeneous Orthogonal Class-Specific Discriminant Analysis}
\newacro{ROCSDA}[ROCSDA]{Regularized Orthogonal Class-Specific Discriminant Analysis}
\newacro{CSSR}[CSSR]{Class-Specific Spectral Regression}
\newacro{NPT}[NPT]{Nonlinear Projection Trick}
\newacro{RBF}[RBF]{Radial Basis Function}
\newacro{SVD}[SVD]{Singular Value Decomposition}
\newacro{SVM}[SVM]{Support Vector Machine}
\newacro{AP}[AP]{Average Precision}
\maketitle


\begin{abstract}
In this paper, we carry out null space analysis for \ac{CSDA} and formulate a number of solutions based on the analysis. We analyze both theoretically and experimentally the significance of each algorithmic step. The innate subspace dimensionality resulting from the proposed solutions is typically quite high and we discuss how the need for further dimensionality reduction changes the situation. 
Experimental evaluation of the proposed solutions shows that the straightforward extension of null space analysis approaches to the class-specific setting can outperform the standard \ac{CSDA} method. Furthermore, by exploiting a recently proposed out-of-class scatter definition encoding the multi-modality of the negative class naturally appearing in class-specific problems, null space projections can lead to a performance comparable to or outperforming the most recent CSDA methods.
\end{abstract}

 \acresetall

\section{Introduction}\label{sec:intro}

\blfootnote{\textcopyright 2019 IEEE.  Personal use of this material is permitted.  Permission from IEEE must be obtained for all other uses, in any current or future media, including reprinting/republishing this material for advertising or promotional purposes, creating new collective works, for resale or redistribution to servers or lists, or reuse of any copyrighted component of this work in other works.} Class-specific discrimination finds application in problems where the objective is to discriminate a class of interest from any other possibility. One of the most notable examples of class-specific problems is person identification, e.g., through face or motion analysis \cite{kittler2000faceverification,iosifidis2012TIFS}. Different from person recognition, which is a multi-class classification problem defined on a pre-defined set of identity classes, person identification discriminates a person of interest from all the other people, i.e., is a binary problem. The application of \ac{LDA} \cite{duda2000pattern,iosifidis2013optimal,iosifidis2014krda} in such binary problems leads to one-dimensional discriminant subspace due to the rank of the adopted between-class scatter matrix. \ac{CSDA} \cite{kittler2000faceverification,goudelis2007classspecific,zafeiriou2012regularized,arashloo2014csksr,iosifidis2015CSRDA,iosifidis2015cskernelspace} allows discriminant subspaces of higher dimensionality by defining suitable intra-class and out-of-class scatter matrices. This leads to better class discrimination in such binary problems compared to \ac{LDA} \cite{goudelis2007classspecific,arashloo2014csksr,iosifidis2015CSRDA}.

Existing methods for \ac{CSDA} typically operate on nonsingular scatter matrices and seek for discriminant directions in the span of the positive training data. In the case where the data dimensionality is higher than the cardinality of the training set, the scatter matrices are singular and regularization is applied to address computational stability problems.
However, experience from multi-class discriminant analysis approaches dealing with this \emph{small sample size problem} \cite{huang2002St} indicates that null space directions contain high discrimination power \cite{chen2000lda,ye2006computational,ye2006null}. Interestingly, one-class discrimination approaches based on null space analysis have been also recently proposed \cite{bodesheim2013kernel,dufrenois2016null}. However, the latter ones are in fact designed by following a multi-class setting and cannot be directly extended for class-specific discrimination.

In this paper, we provide a null space analysis for \ac{CSDA}. We then formulate straightforward class-specific variants of \ac{NLDA} \cite{chen2000lda, huang2002St}, and closely related \ac{ULDA} \cite{jin2001uncorrelated, jin2001face, ye2006reduction}, \ac{OLDA} \cite{ye2005characterization}, and \ac{ROLDA} \cite{ye2006computational} methods. We carry out a detailed evaluation of the significance of each algorithmic step both theoretically and experimentally and discuss different implementation strategies. Furthermore, we combine the concepts of null space analysis with a recently proposed out-of-class scatter definition encoding the multi-modality of the negative class naturally appearing in class-specific problems \cite{iosifidis2018probabilistic} and propose heterogeneous extensions of the class-specific null space algorithms. Our experimental evaluation of the proposed methods shows that the straightforward extensions can outperform the baseline \ac{CSDA} algorithm, while the heterogeneous extensions can achieve a performance comparable to or outperforming the most recent \ac{CSDA} extensions.    

The rest of the paper is organized as follows. In Section~\ref{sec:ProblemStatement}, we give the generic problem statement for class-specific subspace learning. In Section~\ref{sec:CSDA}, we introduce the standard \ac{CSDA} algorithm as well as recent extensions. The main contributions of this paper are described in Section~\ref{sec:nullspace}, where we provide the null space analysis and propose a number of \ac{CSDA} extensions exploiting the analysis. We give our experimental results in Section~\ref{sec:results} and conclude the paper in Section~\ref{sec:conclusions}. 

\section{Problem Statement}\label{sec:ProblemStatement}

Let us denote by $\dataxi \in \R^\noDim, \:i=1,\dots,\noItems$ the training vectors, each followed by a binary label $l_i \in \{1,-1\}$, where a label $l_i = 1$ indicates that sample $i$ belongs to the class of interest, or the positive class, while a label $l_i = -1$ indicates that sample $i$ belongs to the negative class. In practice, the latter case corresponds to a sample belonging to one of the subclasses forming the negative class, the labels of which are not available during training either because these subclasses are not sampled adequately well or because they are expensive to annotate. We want to map the training vectors to a lower-dimensional feature space, i.e., $\dataxi \in \R^\noDim \rightarrow \pdata_i \in \R^\nodim, \nodim \le \noDim$, so that the discrimination of the positive class to the negative class is increased.

A basic assumption in class-specific learning is that the two classes are not linearly separable, but the negative samples lay in multiple directions around the samples of the class of interest. Therefore, class-specific methods typically rely on non-linear approaches. To non-linearly map $\dataxi \in \R^\noDim$ to $\mathbf{\pdata}_i \in \R^\nodim$, traditional kernel-based learning methods map the training vectors to an intermediate feature space $\mathcal{F}$ using a function $\phi(\cdot)$, i.e., $\dataxi \in \R^\noDim \rightarrow \phi(\dataxi) \in \mathcal{F}$. Then, linear class-specific projections are defined by exploiting the Representer theorem and the non-linear mapping is implicitly performed using the kernel function encoding dot products in the feature space, i.e., $\kappa(\dataxi,\datax_j) = \phi(\dataxi)^T \phi(\datax_j)$ \cite{Schlkopf2001}. In this way, a non-linear projection from $\R^\noDim$ to $\R^\nodim$ is expressed as a linear transformation of the kernel matrix $\K \in \R^{\noItems \times \noItems}$ having as elements the pair-wise dot products of the training data representations in $\mathcal{F}$ calculated using the kernel function $\kappa(.,.)$. 

One can also exploit the \ac{NPT} \cite{kwak2013NPT} and apply first an explicit non-linear mapping $\dataxi \in \R^\noDim \rightarrow \data_i \in \R^\noNPTDim, \:\noNPTDim = \rank(\K)$, where $\K \in \R^{\noItems \times \noItems}$ has elements $[\K]_{ij} = \kappa(\dataxi,\datax_j) = \phi(\dataxi)^T \phi(\datax_j)$. This is achieved by setting $\Data = [\data_1,\dots,\data_\noItems] = \mathbf{\Sigma}_\noNPTDim^{\frac{1}{2}} \mathbf{U}_\noNPTDim^T$, where $\mathbf{\Sigma}_\noNPTDim$ and $\mathbf{U}_\noNPTDim$ contain the non-zero eigenvalues and the corresponding eigenvectors of the \emph{centered} kernel matrix $\K$. For an unseen test sample $\datax_l^{test}$, the corresponding mapping is performed as $\data_l^{test} = \mathbf{\Sigma}_\noNPTDim^{-\frac{1}{2}} \mathbf{U}_\noNPTDim^T \mathbf{k}_l^{*test}$, where $\mathbf{k}_l^{*test}$ is the centered version of the (uncentered) kernel vector $\mathbf{k}_l^{test}$ having elements $[\mathbf{k}_l^{test}]_{i}= \phi(\dataxi)^T\phi(\datax_l^{test})$. In the cases, where the size of training set is prohibitive for applying \ac{NPT}, approximate methods for kernel subspace learning can be used, like the one in \cite{iosifidis2016anpt}. After applying \ac{NPT}, a linear projection $\pdata_i = \Proj^T \datai$, where $\Proj \in \R^{\noNPTDim \times \nodim}$, corresponds to a nonlinear mapping from $\R^\noDim$ to $\R^\nodim$. In the rest of this paper, we assume that the data has been preprocessed with \ac{NPT}, which allows to obtain non-linear mappings with linear formulations.

\section{Class-Specific Discriminant Analysis}\label{sec:CSDA}

\ac{CSDA} \cite{goudelis2007classspecific} defines the optimal projection matrix $\Proj$ as the one projecting the data representations of the positive class as close as possible to the positive class mean while at the same time maximizing the scatter of the negative class data from the positive class mean. This objective is achieved by calculating the out-of-class and intra-class scatter matrices w.r.t. the positive class mean $\centroidp \in \R^\noNPTDim$:
\begin{eqnarray*}
\Sn &=& \sum_{l_i = -1} \left( \datai - \centroidp \right)\left( \datai - \centroidp \right)^T  \\
\Sp &=& \sum_{l_i = 1} \left( \datai - \centroidp \right)\left( \datai - \centroidp \right)^T.
\end{eqnarray*}
The optimal $\Proj$ is the one maximizing the following criterion:
\begin{equation}\label{Eq:CSKDA_J}
\mathcal{J}(\Proj) = \trace \left (\left( \Proj^T \Sp \Proj \right)^{-1} \left( \Proj^T \Sn \Proj \right) \right),
\end{equation}
where $\trace(\cdot)$ is the trace operator. $\Proj$ is obtained by solving the generalized eigenproblem $\Sn \proj = \lambda \Sp \proj$ and keeping the eigenvectors in the row the space of both scatter matrices corresponding to the $\nodim$ largest eigenvalues \cite{jia2009trace}, where $\nodim \leq \min(\noItems_p -1, \noDim)$ with the assumption that $\noItems_p \leq \noItems_n$. By defining the total scatter matrix as
\begin{equation*}
\St = \sum_{i = 1}^\noItems \left( \datai - \centroidp \right)\left( \datai - \centroidp \right)^T,  
\end{equation*}
we can easily see that $\St = \Sp + \Sn$ and we get two equivalent optimization criteria for \ac{CSDA}: 
\begin{eqnarray*}
\mathcal{J}_2(\Proj) &=& \trace \left (\left( \Proj^T \Sp \Proj \right)^{-1} \left( \Proj^T \St \Proj \right) \right) \\
\mathcal{J}_3(\Proj) &=& \trace \left (\left( \Proj^T \St \Proj \right)^{-1} \left( \Proj^T \Sn \Proj \right) \right).
\end{eqnarray*}

While the scatter matrices are symmetric (and also the inverse of a symmetric matrix is symmetric), their product is typically not symmetric, which means that the eigenvectors of $(\Sp^{-1})\Sn$ used as the solution of \ac{CSDA} are not guaranteed to be orthogonal. Furthermore, when the data dimensionality leads to rank deficient scatter matrices, the inverse of $\Sp^{-1}$ cannot be directly computed. This is known as the \emph{small sample size problem} \cite{huang2002St}. 

A long line of research has considered these issues for \ac{LDA} and \ac{KDA} including variants such as regularized \ac{LDA} \cite{regularizedLDA}, pseudo-inverse \ac{LDA} \cite{ye2004pseudo}, \ac{NLDA} \cite{chen2000lda, huang2002St}, \ac{ULDA} \cite{jin2001uncorrelated, jin2001face, ye2006reduction}, \ac{OLDA} \cite{ye2005characterization}, and \ac{ROLDA} \cite{ye2006computational}. However, the problems induced by the small sample size problem and the scatter matrix singularity have not received much attention in connection with \ac{CSDA}. A common approach is to follow the approach of regularized \ac{LDA} and solve for the eigenvectors of $(\Sp + \mu\I) ^{-1}\Sn$, where $\mu$ is a small positive value and $\I$ is an identity matrix. The drawback of this approach is the additional hyperparameter $\mu$, the value of which may have a significant impact on the results and is usually determined by following a cross-validation process. 

A Spectral Regression \cite{cai2007spectral} based solution of (\ref{Eq:CSKDA_J}) was proposed in \cite{arashloo2014csksr,iosifidis2015CSRDA}. It has been shown in \cite{iosifidis2016sucskda} that the spectral regression based solution of (\ref{Eq:CSKDA_J}) can be efficiently calculated by exploiting the labeling information of the training vectors. The equivalence of (\ref{Eq:CSKDA_J}) to a low-rank regression problem in which the target vectors can be determined by exploiting the labeling information of the training vectors was proposed in \cite{iosifidis2017cskdaRev}. Finally, a probabilistic framework for class-specific subspace learning was recently proposed in \cite{iosifidis2018probabilistic}, encapsulating criterion (\ref{Eq:CSKDA_J}) as a special case.

\section{Null Space Analysis for CSDA}\label{sec:nullspace}

In the rest of the paper, we assume that the data is centered to the positive class mean. This can always be done by setting $\datai-\centroidp \rightarrow \datai$. Then, the total, intra-class, and out-of-class scatter matrices are given as
\begin{equation*}
\St = \Data \Data^T, \:\:\:\:\Sp = \Data_p \Data_p^T, \:\:\:\: \textrm{and} \:\:\:\: \Sn = \Data_n \Data_n^T,
\end{equation*}
where $\Data_p \in \R^{\noNPTDim \times \noItems_p}$ and $\Data_n \in \R^{\noNPTDim \times \noItems_n}$ with $\noItems_p + \noItems_n = \noItems$ are matrices having as columns the positive and negative training data, respectively. For linearly independent training samples $\datai, \:i=1,\dots,\noItems$, we have $\rank(\Sp) = \min\left(\noItems_p-1, \noNPTDim\right)$, $\rank(\Sn) = \min\left(\noItems_n, \noNPTDim\right)$ and $\rank(\St) = \min \left( \noItems-1, \noNPTDim\right)$. When $\noItems_p -1 \geq \noNPTDim$ and $\noItems_n \geq \noNPTDim$, all scatter matrices are full rank and the corresponding null spaces are empty. As the null spaces are the main focus of this paper, we concentrate on cases, where $\noItems - 1 \leq \noNPTDim$. Thus, we have $\rank(\St) = \noItems-1 = \rank(\Sn) + \rank(\Sp)$.

Here, we should note that linear independence of training vectors $\dataxi, \:i=1,\dots,\noItems$ does not necessarily imply linear independence of $\datai, \:i=1,\dots,\noItems$ for any given kernel function. However, for widely used kernel functions, like the \ac{RBF} and linear kernels, this connection exists. For data representations in $\R^r$ obtained by applying \ac{NPT}, we have $\St$ with full rank.
This is because the dimensions corresponding to the zero eigenvalues of $\K$ have been discarded and because $\St$ and $\K$ are symmetrizable matrix products ($\K = \Phi^T \Phi$ and $\St = \Data \Data^T$) meaning that they share the same eigenvalues \cite{wermuth1993eigenanalysis}. We have $r = \noItems-1$, whenever the training vectors $\dataxi, \:i=1,\dots,\noItems$ are linearly independent and $\noDim \geq \noItems-1$. 
 
To define discriminant directions for our null space CSDA, we will follow ideas similar to those in Null Foley-Sammon transform \cite{guo2006null}. That is, the projection matrix $\Proj$ is formed by the projection directions satisfying
\begin{eqnarray}
\proj^T \Sp \proj &=& 0  \label{Eq:Sp_crit} \\
\proj^T \Sn \proj &>& 0. \label{Eq:Sn_crit}
\end{eqnarray}
The vectors satisfying the above expressions are called null projections and, since $\mathcal{J}(\Proj) \rightarrow \infty$, lead to the best separability of the positive and negative classes. As $\St = \Sp + \Sn$, the projections satisfying both (\ref{Eq:Sp_crit}) and (\ref{Eq:Sn_crit}) satisfy
\begin{equation}
\proj^T \St \proj > 0. \label{Eq:St_crit}
\end{equation}

In order to further analyze the null space projections $\proj$, we denote by $\Sx$ any of the above-defined scatter matrices and define the null and orthogonal complement spaces of a matrix $\Sx$ as follows: $\Null_x = \{ \proj \in \R^\noDim | \Sx \proj = 0\}$ and $\Null_x^{\bot} = \{ \projj \in \R^\noDim | \projj^T\proj = 0 \:\: \forall \: \proj \in \Null_x \}$, respectively. Here, $\Null_x^{\bot}$ is the row space of $\Sx$ (and as $\Sx$ is symmetric it is also equal to the column space). From the above equations, we see that $\proj \in \left(\Nullp \bigcap \Nulln^{\bot}\right) \rightarrow \proj \in \Nullt^{\bot}$. Similarly, $\proj \in \left(\Nulln \bigcap \Nullp^{\bot}\right) \rightarrow \proj \in \Nullt^{\bot}$.  Moreover, $\proj$ is a null projection of $\St$, i.e., $\proj^T \St \proj = 0$, only if $\proj^T \Sp \proj = 0$ and $\proj^T \Sn \proj = 0$, since $\St = \Sp + \Sn$ and all three matrices are positive semi-definite. Thus, we have shown that
\begin{equation}
\Nullt = \Nullp \bigcap \Nulln. 
\label{eq:Nt}
\end{equation}
When $\St$ is full rank, we have $\Nullt = \emptyset$, which means that $\Nullp = \Nulln^{\bot}$, i.e., the directions satisfying (\ref{Eq:Sn_crit}) (or equivalently (\ref{Eq:St_crit})) also satisfy (\ref{Eq:Sp_crit}). This implies also that $\nullity(\Sp) = \rank(\Sn)$ (and $\nullity(\Sn) = \rank(\Sp))$. When $\Nullt = \emptyset$ does not hold, it can be achieved by mapping the data to the row space of $\St$.

The eigenvectors corresponding to zero eigenvalues span the null space of a matrix, while the eigenvectors corresponding to non-zero eigenvalues do not necessarily span the whole row space. However, since we are dealing with symmetric matrices having orthogonal eigenvectors the whole space is spanned. Furthermore, when $\Nullt = \emptyset$, the eigenvalues of $\St$ are the positive eigenvalues of $\Sp$ and $\Sn$, i.e.,
\begin{equation}
\left\{\lambda_{1},...,\lambda_{\nodim}\right\} = \left\{\rho_{1},...,\rho_{\rank(\Sp)}, \nu_{1},...,\nu_{\rank(\Sn)}\right\}, 
\label{eq:eigenvalues}
\end{equation}
where $\St \in \R^{\nodim \times \nodim}$, $\lambda_{1},...,\lambda_{\nodim}$ are the eigenvalues of $\St$, $\rho_{1},...,\rho_{\rank(\Sp)}$ are the non-zero eigenvalues of $\Sp$, and $\nu_{1},...,\nu_{\rank(\Sn)}$ are the non-zero eigenvalues of $\Sn$. This is because for the non-zero eigenvalues of $\Sn$, we have $\Sn  \projj = (\St - \Sp) \projj = \St \projj + 0 = \rho \projj$ and in a similar manner the non-zero eigenvalues of $\Sp$ are eigenvalues of $\St$. As $\rank(\St) = \rank(\Sp) + \rank(\Sn)$, the non-zero eigenvalues of $\Sp$ and $\Sn$ form the full set of eigenvalues of $\St$. However, we observed experimentally that for most datasets $\St$ is quite ill-conditioned (i.e., it has a large ratio of largest to smallest eigenvalues). We also observed numerical instability occurring especially in computations involving eigenvalues of $\Sn$. \eqref{eq:eigenvalues} typically does not hold accurately and the null space of $\Sp$ and the row space of $\Sn$ are not properly aligned.    

Now we proceed to formulate \ac{CSDA} extensions based on the null space analysis. We will analyze the significance of each algorithmic step and also discuss the consequences of the above-mentioned numerical instability.
\vspace{-4pt}

\subsection{Null Space Class-Specific Discriminant Analysis}\label{ssec:NCSDA}

In this section, we propose \ac{NCSDA}, where we exploit similar steps as proposed for the original \ac{NLDA} \cite{chen2000lda} and its extensions \cite{huang2002St, ye2006computational}. We aim at exploiting the discriminant information available in the null space of intra-class scatter matrix $\Sp$ by maximizing the following constrained criterion:
\begin{equation}\label{Eq:NCSDA_J}
\begin{aligned}
&\mathcal{J}_N(\Proj) = &&\trace \left( \Proj^T \Sn \Proj \right) \\
&\mathrm{s.t.} &&\Proj^T \Sp \Proj = 0.
\end{aligned}
\end{equation}

As for \ac{NLDA}, the main idea is to first remove the null space of $\St$ to ensure $\Nullp = \Nulln^\bot$, then map the data to the remaining null space of $\Sp$, and finally maximize $\Sn$ there. A major difference w.r.t. \ac{NLDA} is the final subspace dimensionality innately following from the algorithm. After removing the null space of $\St$, $\nullity(\Sp) = \rank(\Sn)$ and, in the same way, $\nullity(\Sw) = \rank(\Sb)$ for \ac{NLDA}. The rank of the between-class scatter matrix $\Sb$ used in \ac{NLDA} is limited by the number of classes. Thus, the innate subspace dimensionality is low and the \ac{NLDA} methods do not apply any further dimensionality reduction. However, the innate \ac{NCSDA} dimensionality is typically much higher equaling to $\rank(\Sn) = \noItems_n$ and, therefore, it becomes a desired property that the algorithm can, in addition to mapping data to the $\Nullp$, also provide an optimal ranking for the projection vectors so that only some of them can be selected to obtain lower-dimensional final representations.  The pseudo-code of the proposed \ac{NCSDA} is given in Algorithm~\ref{alg:NCSDA} and each step is analyzed below.  
\begin{algorithm}
  \caption{Null space \ac{CSDA}} \label{alg:NCSDA}
  \begin{algorithmic}[1]
		\Statex \textbf{Input:} data matrices $\Data_p$ and $\Data_n$ ($\Data$ = [$\Data_p$, $\Data_n$])
		\Statex \textbf{Output:} projection matrix $\Proj$
		\algrule
		\algcomment{Find row space of $\St$ using reduced \acs{SVD} as in (\ref{Eq:reducedSVD})}
    \State $\Data = \SVDU_t \SVDS_t \SVDV_t^T$ \label{st:spant}		
		\algcomment{Map data to row space of $\St$ and form scatter matrices used in Step \ref{st:nullp}}
	  \State $\tildeData_x = \SVDU_t^T \Data_x$ \:\:\:\: $\tildeSx = \tildeData_x \tildeData_x^T$ \label{st:spant2}		
		\algcomment{Find null space of $\tildeSp$}
		\State Compute $\Projj$ via one of eigenproblems in \eqref{eq:nullSp}-\eqref{eq:nullSnSt} \label{st:nullp}
		\algcomment{Maximize $\tildeSn$ in $\Nullp$ }
		\State (Solve $\Mapp$ from $\Projj^T \tildeSn \Projj \mapp =  \lambda \mapp)^1$ \label{st:spann}
		\algcomment{Combine projection matrices}
		\State $\Proj = \SVDU_t \Projj (\Mapp)^1$ \label{st:combine}
		\algcomment{Orthogonalize $\Proj$ via QR-decomposition}
		\State ($\Proj = \mathbf{QR}$ \:\:\:\: $\Proj = \mathbf{Q})^2$ \label{st:QR}
		\algrule
		\Statex \footnotesize{$(x)$ denotes an optional step $x$}
		\Statex \footnotesize{$(x)^1$ is recommended with \eqref{eq:nullSp} in Step~\ref{st:nullp}}
		\Statex \footnotesize{$(x)^2$ is recommended with \eqref{eq:nullSpSn} or \eqref{eq:nullSnSt} in Step~\ref{st:nullp}}
  \end{algorithmic}
\end{algorithm}

Steps~\ref{st:spant}-\ref{st:spant2} remove the null space of $\St$ to obtain $\Nullp = \Nulln^{\bot}$ (due to (\ref{eq:Nt})). 
We follow \cite{ye2006computational} and use \ac{SVD} to get $\Data= \SVDU \SVDS \SVDV^T$, where $\SVDU$ and $\SVDV$ are orthogonal, 
\begin{equation*}
\SVDS = 
\begin{pmatrix}
\SVDS_t & 0 \\
0 & 0
\end{pmatrix},
\end{equation*}  
$\SVDS_t \in \R^{t \times t}$ is a diagonal matrix with positive diagonal elements in decreasing order and $t = \rank(\St)$. Therefore, 
\begin{equation*}
\begin{aligned}
&\St = \Data \Data^T = \SVDU \SVDS \SVDV^T \SVDV \SVDS^T \SVDU^T = \\
&\SVDU \SVDS \SVDS^T \SVDU^T = \SVDU 
\begin{pmatrix}
\SVDS_t^2 & 0 \\
0 & 0
\end{pmatrix}
\SVDU^T.
\end{aligned}
\end{equation*}
$\SVDU$ can be partitioned as $\SVDU = [\SVDU_t, \SVDU_{t\bot}]$, where $\SVDU_t \in \R^{\noNPTDim \times t}$ contains the eigenvectors of $\St$ corresponding to non-zero eigenvalues,  $\SVDU_{t\bot} \in \R^{\noNPTDim \times (\noNPTDim- t)}$ contains the eigenvectors corresponding to zero eigenvalues, and $\SVDS_r^2$ contains the non-zero eigenvalues. The reduced \ac{SVD} of $\Data$ can be now given as 
\begin{equation}
\Data = \SVDU_t \SVDS_t \SVDV_t^T.
\label{Eq:reducedSVD}
\end{equation}
Step~\ref{st:spant2} projects the data to the subspace spanned by the columns of $\SVDU_t$ to remove the null space of $\St$. We denote the scatter matrices of projected data as $\tildeSt$, $\tildeSp$, and $\tildeSn$. 

We note that Step~\ref{st:spant2} corresponds to applying uncentered \ac{PCA} on the data centered to the positive class mean with the dimensionality set to $t = \rank(\St)$.  As discussed above, for data matrix $\Data$ obtained by applying \ac{NPT}, $\St$ is full rank, i.e., $\Nullt = \emptyset$. In this case, Steps \ref{st:spant} and \ref{st:spant2} are not needed, but we keep them in the algorithm to ensure that $\Nullp = \Nulln^{\bot}$ for any input data.

In Step~\ref{st:nullp}, the null space of $\tildeSp$ is computed. It is the most critical step in \ac{NCSDA}. Following the approach of \cite{huang2002St, guo2006null}, the null space can be found by solving the eigenproblem 
\begin{equation}
\tildeSp  \projj = \lambda \projj
\label{eq:nullSp}
\end{equation}
and forming the projection matrix $\Projj$ from the eigenvectors corresponding to the zero eigenvalues. As $\tildeSp $ is symmetric, the resulting projection vectors will be orthogonal. While these projection vectors span the null space of $\tildeSp$, the zero eigenvalues do not provide any additional information for ranking the vectors when an additional dimensionality reduction is desired. 

\begin{algorithm}
  \caption{Uncorrelated/Orthogonal/Regularized Orthogonal \ac{CSDA}} \label{alg:OCSDA}
  \begin{algorithmic}[1]
		\Statex \textbf{Input:} data matrices $\Data_p$ and $\Data_n$ ($\Data$ = [$\Data_p$, $\Data_n$])
		\Statex \textbf{Output:} projection matrix $\Proj$
		\algrule
		\algcomment{Find row space of $\St$ using reduced \acs{SVD} as in (\ref{Eq:reducedSVD})}
    \State $\Data = \SVDU_t \SVDS_t \SVDV_t^T$ \label{st:spantO}
		\algcomment{Form the projection matrix $\Projjj$}
		\State \acs{UCSDA} and \acs{OCSDA}: $\Projjj = \SVDU_t \SVDS_t^{-1}$
		\Statex \acs{ROCSDA}: $\Projjj = \SVDU_t (\SVDS_t + \alpha\I)^{-1}$
		\algcomment{Map data to row space of $\St$ and form scatter matrices used in Step \ref{st:spannO}}
	  \State $\tildeData_x = \Projjj^T \Data_x$ \:\:\:\: $\tildeSx = \tildeData_x \tildeData_x^T$ \label{st:spant2O}		
		\algcomment{Find $\Projj$ spanning row space of $\tildeSn$}
		\State Compute $\Projj$ via \eqref{eq:svdn}, \eqref{eq:svdp}, or \eqref{eq:nullSnSp}  \label{st:spannO}
		\algcomment{Combine projection matrices}
		\State $\Proj = \Projjj \Projj$ \label{st:combineO}
		\algcomment{Orthogonalize $\Proj$ via QR-decomposition}
		\State \acs{OCSDA} and \acs{ROCSDA}: $\Proj = \mathbf{QR} \:\:\:\: \Proj = \mathbf{Q}$ \label{st:QRO}
  \end{algorithmic}
\end{algorithm}

As $\Nullp = \Nulln^{\bot}$, we can turn our attention to $\tildeSn$ and solve the eigenproblem
\begin{equation}
\tildeSn  \projj = \lambda \projj
\label{eq:nullSn}
\end{equation}
to find the vectors spanning the row space of $\tildeSn$ (i.e, select the eigenvectors corresponding to non-zero eigenvalues). Also this approach results in orthogonal projection vectors and it allows to rank them according to their ability to maximize $\trace \left( \Proj^T \Sn \Proj \right)$ in $\mathcal{J}_N$ \eqref{Eq:NCSDA_J}. However, our experiments show that the projection vectors obtained by solving \eqref{eq:nullSn} fail to span the null space of $\tildeSp$ due to the numerical instability discussed above. This makes the classification performance poor.

Therefore, we also investigate the use of the following generalized eigenproblems in Step~\ref{st:nullp} to analyze their ability to provide null projections for $\tildeSp$ and to rank the projection vectors for further dimensionality reduction: 
\begin{equation}
\tildeSp \projj = \left(\tildeSn + \mu\I\right) \lambda \projj,
\label{eq:nullSpSn}
\end{equation}
\begin{equation}
\tildeSn \projj = \left(\tildeSp + \mu\I\right) \lambda \projj,
\label{eq:nullSnSp}
\end{equation}
\begin{equation}
\tildeSn \projj = \tildeSt \lambda \projj,
\label{eq:nullSnSt}
\end{equation}
where $\mu$ is a small positive value and $\I$ is an identity matrix. To obtain the projection vectors, we select the eigenvectors corresponding to the zero eigenvalues for \eqref{eq:nullSpSn}, but the eigenvectors corresponding to the non-zero eigenvalues for \eqref{eq:nullSnSp} and \eqref{eq:nullSnSt}. Other possible generalized eigenproblems to consider could be $\tildeSp \projj = \tildeSt \lambda \projj$ and $\tildeSt \projj = \left(\tildeSp + \mu\I\right) \lambda \projj$. However, we leave them out, because the analysis for the former combines the elements (drawbacks) of those for \eqref{eq:nullSpSn} and \eqref{eq:nullSnSt}, while the latter gives exactly the same results as \eqref{eq:nullSnSp} when selecting the eigenvalues smaller than one.

The projection vectors resulting from the generalized eigenproblems are no longer guaranteed to be orthogonal. However, for symmetric matrices $\mathbf{A}$ and $\mathbf{B}$ and for a positive definite $\mathbf{B}$, the generalized eigenproblem $\mathbf{A} \projj = \mathbf{B} \lambda \projj$ has real eigenvalues and the eigenvectors are linearly independent and $\mathbf{B}$-orthogonal, i.e., $\projj_i^T \mathbf{B} \projj_j = 0$ for $i \neq j$ \cite{Parlett1998symmetric}. All the scatter matrices are symmetric and positive semi-definite. After removing the null space of $\St$, $\tildeSt$ is full rank and, therefore, positive definite. Furthermore, the regularization applied in \eqref{eq:nullSpSn} and \eqref{eq:nullSnSp} preserves the symmetry, while making $\tildeSn$ and $\tildeSp$ positive definite. Thus, \eqref{eq:nullSpSn}, \eqref{eq:nullSnSp}, and \eqref{eq:nullSnSt} will have  $\tildeSn$-orthogonal, $\tildeSp$-orthogonal, and $\tildeSt$-orthogonal eigenvectors, respectively. For \eqref{eq:nullSnSp} and \eqref{eq:nullSnSt}, the linear independence of the eigenvectors is important to maintain the assumption that the eigenvectors corresponding to the non-zero eigenvalues span the row space of $\tildeSn$ and, thus, the null space of $\tildeSp$ due to \eqref{eq:Nt}. For \eqref{eq:nullSnSp}, we have $\Projj^T\left(\tildeSp + \mu\I\right)\Projj = \I$. Since we select only the eigenvectors spanning the row space of $\tildeSn$ to form $\Projj$ and assume that they are null projections for $\tildeSp$, we get $\Projj^T\left(\tildeSp + \mu\I\right)\Projj = \mu \Projj^T \Projj = \I \rightarrow  \Projj^T\Projj = \frac{1}{\mu} \I$, i.e., the projection vectors are orthogonal. 

Considering the usefulness of \eqref{eq:nullSpSn}-\eqref{eq:nullSnSt} for projection vector ranking, the eigenvectors corresponding to zero eigenvalues are used in  
\eqref{eq:nullSpSn} and, therefore, the eigenvalues do not offer ranking information. Furthermore, after mapping the data to the row space of $\St$ all the non-zero generalized eigenvalues computed with respect to $\tildeSt$ as in \eqref{eq:nullSnSt} are equal to one and, thus, they are also useless for ranking. Only for \eqref{eq:nullSnSp} we have non-zero eigenvalues which can be directly used for ranking the projection vectors. We also note that \eqref{eq:nullSnSp} corresponds to applying the standard \ac{CSDA} in the row space of $\St$, which results in orthogonal projection vectors. In our experiments, the projection vectors solved from \eqref{eq:nullSpSn}-\eqref{eq:nullSnSt} span the null space of $\tildeSp$ with same accuracy as the solution of \eqref{eq:nullSp}.

Step~\ref{st:spann} aims at finding a mapping $\Mapp$ that maximizes $\tildeSn$ in the null space of $\tildeSp$. The mapping can be formed by solving the eigenproblem 
$\Projj^T \tildeSn \Projj \mapp =  \lambda \mapp$
and taking the eigenvectors corresponding to non-zero eigenvalues. A corresponding step was a part of the original \ac{NLDA} \cite{chen2000lda}. However, it was considered unnecessary in \cite{huang2002St} since $\tildeSn$, or $\Sb$ in the case of \ac{NLDA}, has no null space to remove ($\tildeSn$ should be full rank after removing the null space of $\St$ and mapping to null space of $\tildeSp$) and 
it was proved in \cite{ye2006computational} that the projection $\Mapp$ has now effect in the algorithm. This is because for any orthogonal matrix $\mathbf{A}$, $\trace(\mathbf{A}^T \mathbf{B} \mathbf{A}) = \trace(\mathbf{B})$. We know that $\Mapp$ is orthogonal because $\Projj^T \tildeSn \Projj$ is symmetric. Therefore, 
\begin{equation*}
\trace \left( \Mapp^T \Projj^T \SVDU_t^T \Sn \SVDU_t \Projj \Mapp \right) = \trace \left( \Projj^T \SVDU_t^T  \Sn \SVDU_t \Projj \right)
\label{eq:trace}
\end{equation*}
and $\Mapp$ has no effect in maximizing \eqref{Eq:NCSDA_J}. However, these arguments against Step~\ref{st:spann} do not take into account the need for ranking the projection vectors for further dimensionality reduction, while it offers another approach for evaluating the usefulness of the vectors. 

Our experiments confirm that Step~{\ref{st:spann} improves the results when combined with using \eqref{eq:nullSp} in Step~\ref{st:nullp} and the subspace dimension is cut to 1-25. With \eqref{eq:nullSn} and \eqref{eq:nullSnSp}, the vectors have been already ranked to maximize \eqref{Eq:NCSDA_J} and Step~{\ref{st:spann} does not change the results. For \eqref{eq:nullSpSn} and \eqref{eq:nullSnSt}, $\tildeSn$/$\tildeSt$-orthogonality of projection vectors in $\Projj$ is now problematic. For \eqref{eq:nullSpSn}, we have $\Projj^T\left(\tildeSn + \mu\I\right)\Projj \approx \Projj^T\tildeSn\Projj = \I$ and, for \eqref{eq:nullSnSt}, $\Projj^T\tildeSt \Projj  = \Projj^T\tildeSn \Projj + \Projj^T\tildeSp \Projj = \I$, which for the null space of $\tildeSp$ becomes $\Projj^T\tildeSn\Projj = \I$. Thus, in both cases all the eigenvalues in Step~\ref{st:spann} will be equal to one and no further ranking information is gained. 

In Step~\ref{st:combine}, the separate projection matrices are combined to form $\Proj$. Finally, we add an optional step to orthogonalize $\Proj$. This Step~\ref{st:QR} compensates for the lack of orthogonality following from using \eqref{eq:nullSpSn} or \eqref{eq:nullSnSt} in Step~\ref{st:nullp}. When \eqref{eq:nullSp}, \eqref{eq:nullSn}, or \eqref{eq:nullSnSp} is used, the projection vectors are already orthogonal and Step~\ref{st:QR} has no effect.

\vspace{-2pt}
\subsection{Orthogonal CSDA}\label{ssec:OCSDA}

Next, we formulate three modified optimization criteria as straightforward class-specific versions of those originally used for (generalized) \ac{ULDA} \cite{ye2005characterization}, \ac{OLDA} \cite{ye2005characterization}, and \ac{ROLDA} \cite{ye2006computational}. The criterion for the proposed \ac{UCSDA} is
\begin{equation}\label{Eq:UCSDA_J}
\begin{aligned}
&\mathcal{J_U}(\Proj) = &&\trace \left( (\Proj^T \St \Proj)^{+}(\Proj^T \Sn \Proj) \right) \\
&\mathrm{s.t.} &&\Proj \in \R^{r \times \nodim} \:\:\:\: \Proj^T \St \Proj = \I_\nodim,
\end{aligned}
\end{equation}
where $()^{+}$ denotes the pseudo-inverse and $\nodim = \rank(\Sn)$ is the subspace dimensionality. The projection vectors are required to be $\St$-orthogonal, which guarantees that the vectors mapped to the projection space are mutually uncorrelated.  The criterion for the proposed \ac{OCSDA} is almost the same, but the constraint now requires standard orthogonality:
\begin{equation}\label{Eq:OCSDA_J}
\begin{aligned}
&\mathcal{J}_O(\Proj) = &&\trace \left( (\Proj^T \St \Proj)^{+}(\Proj^T \Sn \Proj) \right) \\
&\mathrm{s.t.} &&\Proj \in \R^{r \times \nodim} \:\:\:\: \Proj^T \Proj = \I_\nodim,
\end{aligned}
\end{equation}
Finally, the criterion for \ac{ROCSDA} regularizes the total scatter matrix in $\mathcal{J}_O(\Proj)$: 
\begin{equation}\label{Eq:ROCSDA_J}
\begin{aligned}
&\mathcal{J}_R(\Proj) = &&\trace \left( (\Proj^T (\St + \alpha\I) \Proj)^{+}(\Proj^T \Sn \Proj) \right) \\
&\mathrm{s.t.} &&\Proj \in \R^{r \times \nodim} \:\:\:\: \Proj^T \Proj = \I_\nodim.
\end{aligned}
\end{equation}

Let us consider the criterion in $\mathcal{J}_U(\Proj)$ and $\mathcal{J}_O(\Proj)$ without taking the constraints into account first. We have
\begin{equation*}
\begin{split}
&\trace \left( (\Proj^T \St \Proj)^{+}(\Proj^T \Sn \Proj) \right) \\
&=\trace \left( (\Proj^T \St \Proj)^{+}(\Proj^T \St \Proj) \right) - \trace \left( (\Proj^T \St \Proj)^{+}(\Proj^T \Sp \Proj) \right) \\
&=\rank(\Proj^T \St \Proj) - \trace \left( (\Proj^T \St \Proj)^{+}(\Proj^T \Sp \Proj) \right) \leq \nodim - 0, 
\end{split}
\end{equation*}
where $\nodim$ is the subspace dimensionality, the first equality follows from $\St = \Sp + \Sn$ and basic properties of trace,  the second equality follows from $\trace(\mathbf{A}^{+}\mathbf{A}) = \rank(\mathbf{A})$ for all square matrices, and the inequality is due to $\Proj^T \St \Proj = \tildeSt \in \R^{\nodim \times \nodim} \rightarrow \rank(\tildeSt) \leq \nodim$ and $\trace \left( (\Proj^T \St \Proj)^{+}(\Proj^T \Sp \Proj) \right) \geq 0$ for the positive semi-definite $\Sp$. Thus, the criterion gets its maximum value for a given $\nodim$ if $\tildeSt$ is full rank and $\Proj$ is a null projection for $\Sp$. Therefore, \ac{NCSDA} maximizes the criterion for $\nodim = \rank(\Sn)$ if $\Nullp = \Nulln^{\bot}$. From our null space analysis, we know that this can be obtained by removing the null space of $\St$ if $\rank(\St) = \rank(\Sp)+\rank(\Sn)$. As we consider in this paper only cases where $\noItems - 1 \leq \noNPTDim$, the solution of \ac{NCSDA} always maximizes the criterion and provides a solution to \ac{UCSDA}/\ac{OCSDA} whenever the constraints are satisfied. In Section~\ref{ssec:NCSDA}, we presented solutions which are either $\St$-orthogonal, orthogonal, or both.  
   
The original solutions to generalized \ac{ULDA} \cite{ye2005characterization} and \ac{OLDA} \cite{ye2005characterization} were derived using simultaneous diagonalization of the three scatter matrices. This solution does not require $\rank(\St) = \rank(\Sp)+\rank(\Sn)$ to hold. It can be shown that $\Proj$ simultaneously diagonalizes $\St$, $\Sp$, and $\Sn$ and is a solution to $\mathcal{J_U}(\Proj)$ for $\nodim = \rank(\Sn)$ (see \cite{ye2005characterization, ye2006reduction} for details), when
$\Proj = \SVDU_t \SVDS_t^{-1} \Projj$,
where $\SVDU_t$ and $\SVDS_t$ are as defined in the reduced \ac{SVD} of $\Data$ in \eqref{Eq:reducedSVD} and $\Projj$ is obtained by first mapping the negative samples as $\tildeData_n = \SVDS_t^{-1} \SVDU_t^T \Data_n$ and then applying the reduced \ac{SVD} on $\tildeData_n$ as 
\begin{equation}
\tildeData_n = \SVDUU_q \SVDSS_q \SVDVV_q \:\:\:\: \Projj = \SVDUU_q,
\label{eq:svdn}
\end{equation}
where $\SVDUU_q$ is orthogonal and contains the eigenvectors of $\tildeSn$ corresponding to the non-zero eigenvectors. Thus, $\Projj$ spans the row space of $\tildeSn$.
We now provide a pseudo-code for the proposed \ac{UCSDA}, \ac{OCSDA}, and \ac{ROCSDA} based on the above derivation in Algorithm~\ref{alg:OCSDA}. 

While Steps~\ref{st:spant}-\ref{st:spant2} in Algorithm~\ref{alg:NCSDA} apply uncentered \ac{PCA} on the data, Steps~\ref{st:spantO}-\ref{st:spant2O} in Algorithm~\ref{alg:OCSDA} apply uncentered \ac{PCA} whitening.  In our \ac{NCSDA} experiments, we observed a discrepancy between the vectors spanning the null space of $\tildeSp$ (as solved from \eqref{eq:nullSp}) and vectors spanning the row space of $\tildeSn$ (as solved from \eqref{eq:nullSn}), which we believe to be related with the ill-conditioned total scatter matrix $\tildeSt$. The whitening operation gives $\St$-orthogonal projection vectors and, after mapping, all the eigenvalues of $\tildeSt$ will be ones. Thus, $\tildeSt$ is no longer ill-conditioned. Indeed, we observe the whitening to cure the discrepancy and the projection vectors solved using \eqref{eq:nullSn} become able to satisfy the null constraint for $\Sp$. 

However, due to \eqref{eq:eigenvalues}, all the non-zero eigenvalues of $\tildeSn$ and $\tildeSt$ will be also ones. This removes the ability of the non-zero eigenvalues to rank the projection vectors for dimensionality reduction. Applying \ac{ROCSDA} provides a compromise between the ill-conditioned total scatter matrix and losing the ranking ability due to equalized eigenvalues. The regularized whitening leads to stable eigenvalues close to one, but leaves some variability to use for ranking the projection vectors, which significantly improves the performance. While \ac{ROCSDA} can also prevent numerical errors due to division with very small numbers, this is not relevant for our implementation, because typically a threshold $\epsilon > 0$ (e.g., $\epsilon = 10^{-6}$) is used to decide whether an eigenvalue equals to zero and setting values of $\alpha$ smaller than $\epsilon$ brings the same improvement. We believe that the whitening applied in \ac{UCSDA} and \ac{OCSDA} prevents numerical errors of \ac{NCSDA}, while the main contribution of \ac{ROCSDA} is to improve the projection vector ranking. 
 
Step~\ref{st:spannO} finds vectors spanning the row space of $\tildeSn$. We know from our null space analysis that due to \eqref{eq:Nt} this corresponds to finding the vectors spanning the null space of $\tildeSp$. In fact, we could use any of the eigenproblems used in Step~\ref{st:nullp} of Algorithm~\ref{alg:NCSDA} also here. However, we now confine ourselves to the best performing eigenproblem \eqref{eq:nullSnSp} along with \eqref{eq:svdn} and a similar \ac{SVD} approach for $\tildeData_p$ given as 
\begin{equation}
\tildeData_p = \SVDUU \SVDSS \SVDVV \:\:\:\: \SVDUU = [\SVDUU_q, \SVDUU_{q\bot}] \:\:\:\: \Projj = \SVDUU_{q\bot},
\label{eq:svdp}
\end{equation}
where we select as our projection vectors the columns of $\SVDUU_{q\bot}$, which correspond to the zero singular values in the full \ac{SVD}. 

\vspace{-2pt}
\subsection{Heterogeneous Null Space \ac{CSDA} and Heterogeneous Orthogonal \ac{CSDA}}}\label{ssec:HNCSDA}

Up to this point, we have aimed at maximizing $\Sn$ following the standard \ac{CSDA} assumption, i.e., the negative samples are evenly spread out around the positive class. However, this is typically not the case, but the negative class actually consist of more than one distinct classes. While in class-specific approaches, we assume not to know labels for such classes, we can still cluster the negative data and reformulate the solution so that we allow the clusters of similar items to stay close to each other and concentrate on maximizing the distance of these clusters to the positive class mean. Such an approach has been recently proposed in \cite{iosifidis2018probabilistic} and here we combine the \emph{heterogeneous} formulation for the negative class with the concepts based on our null space analysis.  

We now assume the negative class to be formed of $K$ clusters. The centroid of the $k^{th}$ cluster can be computed as $\centroidk = \frac{1}{\noItems_k} \sum_{j=1}^{\noItems_k} \datakj$, where $\noItems_k$ is the number of items in the cluster and $\datakj$ is the $j^{th}$ sample in the cluster. The out-of-class scatter was earlier defined as $\Sn = \sum_{l_i = -1} \left( \datai - \centroidp \right)\left( \datai - \centroidp \right)^T = \sum_{k=1}^K\sum_{j=1}^{\noItems_k}\left( \datakj - \centroidp \right)^2$. Let us define the negative class within-cluster scatter $\Snw$ and the between-cluster scatter $\Snb$ as  
\begin{eqnarray*}
\Snw &=& \sum_{k=1}^K\sum_{j=1}^{\noItems_k} \left( \datakj - \centroidk \right)\left( \datakj - \centroidk \right)^T\\
\Snb &=& \sum_{k=1}^K \noItems_k \left( \centroidk - \centroidp \right) \left( \centroidk - \centroidp \right)^T.
\end{eqnarray*}
We can show that $\Sn = \Snw + \Snb$ as follows
\begin{equation*}
\begin{aligned}
 \left( \datakj - \centroidp \right) =& \left( \centroidk - \centroidp \right) +  \left( \datakj - \centroidk \right) \Rightarrow\\ 
 \left( \datakj - \centroidp \right)^2 =& \left( \centroidk - \centroidp \right)^2 +  \left( \datakj - \centroidk \right)^2\\ 
&+ 2\left( \centroidk - \centroidp \right)\left( \datakj - \centroidk \right)^T \Rightarrow\\
 \sum_{k=1}^K\sum_{j=1}^{\noItems_k}\left( \datakj - \centroidp \right)^2 =& \sum_{k=1}^K \noItems_k \left( \centroidk - \centroidp \right)^2 + \\
\sum_{k=1}^K\sum_{j=1}^{\noItems_k} \left( \datakj - \centroidk \right)^2 
 + 2 &\sum_{k=1}^K \sum_{j=1}^{\noItems_k}\left( \centroidk - \centroidp \right)\left( \datakj - \centroidk \right)^T \Rightarrow\\
 \Sn =& \Snb + \Snw + 0,\\
\end{aligned}
\end{equation*}
where the zero comes from $\sum_{j=1}^{\noItems_k}\left( \datakj - \centroidk \right) = \sum_{j=1}^{\noItems_k} \datakj - \sum_{j=1}^{\noItems_k} \centroidk
= \noItems \centroidk - \noItems \centroidk$. 

\begin{algorithm}
  \caption{Heterogeneous Null space \ac{CSDA}} \label{alg:HNCSDA}
  \begin{algorithmic}[1]
		\Statex \textbf{Input:} data matrices $\Data_p$ and $\Data_n$ ($\Data$ = [$\Data_p$, $\Data_n$])
		\Statex \textbf{Output:} projection matrix $\Proj$
		\algrule
		\algcomment{Find row space of $\St$ using reduced \acs{SVD} as in (\ref{Eq:reducedSVD})}
    \State $\Data = \SVDU_t \SVDS_t \SVDV_t^T$ \label{st:spantHN}		
		\algcomment{Map data to row space of $\St$ and form scatter matrices used in Step \ref{st:nullpHN}}
	  \State $\tildeData_x = \SVDU_t^T \Data_x$ \:\:\:\: $\tildeSx = \tildeData_x \tildeData_x^T$ \label{st:spant2HN}		
		\algcomment{Find null space of $\tildeSp$}
		\State Solve $\Projj$ from $\tildeSp \projj = \left(\tildeSn + \mu\I\right) \lambda \projj$ \label{st:nullpHN}
		\algcomment{Map negative class to null space of $\tildeSp$}
	  \State $\Data^*_n = \Projj^T \tildeData_n$ 
		\algcomment{Cluster negative samples}
		\State Apply K-means on $\Data^*_n$ to find $\centroidk^*$ for $k=1,...,K$ \label{st:clusterHN}	
		\algcomment{Compute between-cluster scatter matrix $\Snb^*$}
		\State $\Snb^* = \sum_{k=1}^K \noItems_k \left( \centroidk^* - \centroidp^* \right) \left( \centroidk^* - \centroidp^* \right)^T$
		\algcomment{Maximize $\Snb^*$}
		\State Solve $\Mapp$ from $\Snb^* \mapp = \lambda \mapp$ \label{st:spannHN}
		\algcomment{Combine projection matrices}
		\State $\Proj = \SVDU_t \Projj \Mapp$ \label{st:combineHS}
		\algcomment{Orthogonalize $\Proj$ via QR-decomposition}
		\State $\Proj = \mathbf{QR}$ \:\:\:\: $\Proj = \mathbf{Q}$ \label{st:QRHN}
  \end{algorithmic}
\end{algorithm}

We now proceed to propose optimization criteria for two heterogeneous \ac{CSDA} variants exploiting the null space analysis, namely \acf{HNCSDA} and \acf{HOCSDA}.
The criterion for \ac{HNCSDA} is
\begin{equation}\label{Eq:HNCSDA_J}
\begin{aligned}
&\mathcal{J}_{HN}(\Proj) = &&\trace \left( \Proj^T \Snb \Proj \right) \\
&\mathrm{s.t.} && \Proj^T \Sp \Proj = 0
\end{aligned}
\end{equation} 
and for \ac{HOCSDA} 
\begin{equation}\label{Eq:HOCSDA_J}
\begin{aligned}
&\mathcal{J}_{HO}(\Proj) = &&\trace \left( (\Proj^T \St \Proj)^{+}(\Proj^T \Snb \Proj) \right) \\
&\mathrm{s.t.} &&\Proj \in \R^{r \times \noClust} \:\:\:\: \Proj^T \Proj = \I_\nodim.
\end{aligned}
\end{equation}
Maximizing $\trace \left( \Proj^T \Snb \Proj \right)$ instead of $\trace \left( \Proj^T \Sn \Proj \right)$ will try to push the clusters of the negative class far away from the positive class mean while allowing the samples within the clusters be close to each other. This in many cases describes the negative class in a more natural way.   Furthermore, we can see that $\rank(\Snw) = \noItems_n - \noClust$, $\rank(\Snb) = \noClust$, and $\rank(\Sw) = \rank(\Snb) + \rank(\Snb)$ for linearly independent samples. Thus, the innate dimensionality of the proposed heterogeneous methods will be limited by the number of clusters in the negative class, which is low enough to be used as the final subspace dimensionality. The pseudo-code for the proposed \ac{HNCSDA} is presented in Algorithm~\ref{alg:HNCSDA} and the pseudo-code for the proposed \ac{HOCSDA} in Algorithm~\ref{alg:HOCSDA}.

\begin{algorithm}
  \caption{Heterogeneous Orthogonal \ac{CSDA}} \label{alg:HOCSDA}
  \begin{algorithmic}[1]
		\Statex \textbf{Input:} data matrices $\Data_p$ and $\Data_n$ ($\Data$ = [$\Data_p$, $\Data_n$])
		\Statex \textbf{Output:} projection matrix $\Proj$
		\algrule
		\algcomment{Find row space of $\St$ using reduced \acs{SVD} as in (\ref{Eq:reducedSVD})}
    \State $\Data = \SVDU_t \SVDS_t \SVDV_t^T \:\:\:\: \Projjj = \SVDU_t \SVDS_t^{-1}$ \label{st:spantHO}		
		\algcomment{Map negative class to row space of $\tildeSt$}
	  \State $\Data^*_n = \Projjj^T \tildeData_n$ 
		\algcomment{Cluster negative samples}
		\State Apply K-means on $\Data^*_n$ to find $\centroidk^*$ for $k=1,...,K$ \label{st:clusterHO}	
		\algcomment{Compute between-cluster scatter matrix $\Snb^*$}
		\State $\Snb^* = \sum_{k=1}^K \noItems_k \left( \centroidk^* - \centroidp^* \right) \left( \centroidk^* - \centroidp^* \right)^T$
		\algcomment{Maximize $\Snb^*$}
		\State Solve $\Mapp$ from $\Snb^* \mapp = \lambda \mapp$ \label{st:spannHO}
		\algcomment{Combine projection matrices}
		\State $\Proj = \Projjj \Mapp$ \label{st:combineHO}
		\algcomment{Orthogonalize $\Proj$ via QR-decomposition}
		\State $\Proj = \mathbf{QR}$ \:\:\:\: $\Proj = \mathbf{Q}$ \label{st:QRHO}
  \end{algorithmic}
\end{algorithm}

For Step~\ref{st:nullpHN} in \ac{HNCSDA}, we select the generalized eigenproblem $\tildeSp \projj = \left(\tildeSn + \mu\I\right) \lambda \projj$ \eqref{eq:nullSpSn}. For \ac{NCSDA}, it demonstrated a good ability to produce null projections for $\tildeSp$ also for the test samples.  In \ac{HOCSDA}, we do not apply further dimensionality reduction and, therefore, we do not use the regularized whitening of \ac{ROCSDA}. We perform the clustering of the negative samples using the K-means algorithm, where the number of clusters, $K$, is determined via cross-validation. 

Following an analysis similar to one presented in Section~\ref{sec:nullspace}, we can show that $\Nulln = \Nullnw \bigcap \Nullnb$ 
and that the non-zero eigenvalues $\nu_{i}$ of $\Sn$ are equal to the non-zero eigenvalues of $\Snw$ and $\Snb$:
\begin{equation*}
\left\{\nu_{1},...,\nu_{\rank(\Sn)}\right\} = \left\{\theta_{1},...,\theta_{\rank(\Snw)}, \gamma_{1},...,\gamma_{\rank(\Snb)}\right\}, 
\end{equation*}
where $\nu_{1},...,\nu_{\rank(\Sn)}$ are the positive eigenvalues of $\Sn$, $\theta_{1},...,\theta_{\rank(\Snw)}$ are the non-zero eigenvalues of $\Snw$, and $\gamma_{1},...,\gamma_{\rank(\Snb)}$ are the non-zero eigenvalues of $\Snb$. In the proposed heterogeneous approaches, the non-zero eigenvalues of $\Sn$ will be equal to one either due to mapping to the row space of $\Sn$ (\ac{HNCSDA}) or due to whitening the data (\ac{HOCSDA}). However, this is no longer a problem as we do not need to rank the projection vectors for further dimensionality reduction after solving for $\Mapp$ in Step~\ref{st:spannHN} in \ac{HNCSDA} and Step~\ref{st:spannHO} in \ac{HOCSDA}.
  	
Moreover, we see that the direction maximizing $\Snb$ are in the null space of $\Snw$. Thus, the proposed \ac{HNCSDA} method optimizes also a more tightly constrained criterion 
\begin{equation}\label{Eq:HNCSDA_J2}
\begin{aligned}
&\mathcal{J}_{HN2}(\Proj) = &&\trace \left( \Proj^T \Snb \Proj \right) \\
&\mathrm{s.t.} &&\Proj^T (\Sp + \Snw) \Proj = 0.
\end{aligned}
\end{equation} 
Now considering the case $\noClust=1$, we get 
\begin{equation*}
\begin{aligned}
&\Sp + \Snw \\
&= \sum_{l_i = 1} \left( \datai - \centroidp \right)\left( \datai - \centroidp \right)^T + \sum_{l_i = -1} \left( \datai - \centroidn \right)\left( \datai - \centroidn \right)^T \\
&= \sum_{c = 1}^2\sum_{j = 1}^{\noItems_c} \left( \datacj - \centroidc \right)\left( \datacj - \centroidc \right)^T = \Sw*, \\
&\Snb = \noItems_n \left( \centroidn - \centroidp \right) \left( \centroidn - \centroidp \right)^T = \Sb*,\\
\end{aligned}
\end{equation*} 
where $\Sw*$ and $\Sb*$ are the within-class and between-class scatter matrices for the binary \ac{LDA} (with the difference of a scaling factor) and the proposed method is equal to the \ac{NLDA}. On the other hand, setting $\noClust=\noItems_n$ we get		
\begin{eqnarray*}
\Snb &=& \sum_{l_i = -1} \left( \datai - \centroidp \right) \left( \datai - \centroidp \right)^T = \Sn.
\end{eqnarray*}
This shows that the parameter $\noClust$ in the proposed \ac{HNCSDA} in fact acts as a decision parameter to select between \ac{NLDA} and \ac{NCSDA}.

\section{Experiments}\label{sec:results}

\subsection{Datasets}\label{SS:Datasets}

We run our experiments on the following publicly available widely used datasets: \emph{JAFFE} \cite{DSJaffe}, \emph{Cohn-Kanade} \cite{DSKanade}, and \emph{BU} \cite{DSBu}, \emph{ORL} \cite{DSOrl}, \emph{AR} \cite{DSAR}, and \emph{Extended Yale-B} \cite{DSYale}, and 15 scenes \cite{DS15scenes}. JAFFE, Cohn-Kanade, and BU are facial expression datasets having seven different expression classes: angry, disgusted, afraid, happy, sad, surprised, and neutral. JAFFE depicts ten Japanese females each with three images for each expression. Cohn-Kanade depicts 210 persons of age between 18 and 50. We use 35 randomly selected images for each expression class. BU depicts over 100 persons. All expression classes except neutral have four different intensity levels. In our experiments, we use mainly images with the most expressive intensity level. ORL, AR, and Extended Yale-B are face recognition datasets. ORL depicts 40 persons with 10 varying images for each. To focus on small sample size problems (where $\noItems - 1 \leq \noDim$), we pick only the first 15 persons from AR and Extended Yale-B datasets. AR has 26 images of each person. Yale database depicts each person in 64 different illumination conditions. Finally, 15 scenes dataset consists of images from 15 different scenes (such as forest, coast, industrial) having 200-400 images for each scene. 

For the first six datasets, we use 70 \% of the samples for training (further divided for 80\% training and 20\% validation sets in 5-fold cross-validation when needed) and the remaining 30\% for testing. For 15 scenes dataset, we use only 10\% of the samples for training to maintain the small sample set condition. The original feature dimension, the number of samples and classes used is our experiments, and the percentage of images used for training for each dataset are listed in Table~\ref{tab:datasets}. Note that due to the applied \ac{NPT} mapping the methods are run on training vectors having the dimensionality $\noItems - 1$, where $\noItems$ is the number of training samples.

\begin{table}[ht]
\caption{Summary of dataset properties}
\addtolength{\tabcolsep}{-3pt}  
\normalsize
\label{tab:datasets}
\begin{tabular}{l|cccc}
Name & \noDim & \# samples & \# classes & Training \% \\
\hline
JAFFE \cite{DSJaffe} & 1200 & 210 & 7 & 70\%  \\
Kanade \cite{DSKanade} & 1200 & 245 & 7 & 70\%  \\
BU \cite{DSBu} & 1200 & 700 & 7 & 70\%  \\
ORL \cite{DSOrl} & 1200 & 400 & 40 & 70\%  \\
AR \cite{DSAR} & 1200 & 390 & 15 & 70\%  \\
Yale \cite{DSYale} & 1200 & 960 & 15  & 70\%  \\
15 scenes \cite{DS15scenes} &  512 & 4485 & 15 & 10\%  
\vspace{-7pt}
\end{tabular}
\addtolength{\tabcolsep}{3pt}  
\end{table}

\subsection{Experimental Setup}\label{SS:Setup}

The \ac{CSDA} methods are feature reduction methods that map the feature vectors to a lower and more discriminant feature space. Any classifier, such as \ac{SVM}, could be applied on the mapped features to evaluate their discrimination power. However, a common approach with class-specific disriminant methods is to determine the class-specific feature space and compute the positive class centroid, $\centroidp$, during training. The test samples are then mapped to the determined feature space and their similarity to $\centroidp$ is measured. Test samples are ranked according to their similarities and the performance of the algorithms is evaluated using \ac{AP} as for a retrieval method.  We also followed this approach and report the 11-point interpolated \ac{AP} values \cite{Everingham2010Pascal}. 

We preprocessed the data vectors by mapping them to a kernel space using the \ac{NPT} approach as described in Section~\ref{sec:ProblemStatement}. Here we use the \ac{RBF} kernel with $\sigma = \sqrt{ \frac{1}{\noItems*\noDim} \sum_{i=1}^{\noItems} \sum_{j=1}^{\noDim}\data_{ij}}$, where $\noItems$ is the number of training samples and $\data_{ij}$ is the $j^{th}$ element of the $i^{th}$ training vector. Whenever we applied regularization on a scatter matrix (i.e., $\Sx^* = \Sx + \mu*\I$), we set $\mu = 10^{-4}$. In finding zero/non-zero eigenvalues, we considered values smaller than $\epsilon = 10^{-6}$ to be zero. We set the regularization parameter of $\alpha$ in \ac{ROCSDA} to $10^{-7}$  in order to demonstrate that a value smaller than $\epsilon$ works well. The hyperparameters $\nodim$ and $\noClust$ were set by 5-fold cross-validation, where we considered values 1-25 for $\nodim$ and $\noClust$ was selected from $\{1,2,3,5,10\}$. For clustering, we applied the K-means algorithm. Each time, we ran K-means for 10 times and selected the clustering that resulted in the lowest sum of distances between the items and the corresponding centroids.   

We repeated the experiments considering each class in turn as the positive class. The samples belonging to all the other classes were considered as negative. For each positive class, we repeated the whole experiment for five times and the reported results are averages over all the classes and all the repetitions. While the methods themselves do not have random elements, differences arise from the random splitting of the data into train and test sets and the use of the K-means clustering. Nevertheless, all the methods were evaluated with the same splittings. All the experiments were run on Matlab R2017a and the codes will be made publicly available upon publication.


\subsection{Experimental Results}\label{ssec:results}

\subsubsection{Experiments on Null space \ac{CSDA}} \label{sssec:nullexperiments}

We concentrate first on \ac{NCSDA} discussed in Section~\ref{ssec:NCSDA}. We compared different eigenproblems for Step~\ref{st:nullp} in Algorithm~\ref{alg:NCSDA} both with full dimensionality, i.e., $\nodim = \rank(\Sn$), and with a reduced dimensionality, where the best dimensionality within 1-25 was determined by cross-validation as explained above. We also evaluated the optional steps \ref{st:spann} and \ref{st:QR} of Algorithm~\ref{alg:NCSDA} in each setup. The classification results are given in Table~\ref{tab:nullresults}. To gain further insight, we also experimented on the constraint and criterion values for $\mathcal{J}_N$ \eqref{Eq:NCSDA_J}. In Table~\ref{tab:nullcrit}, we report how well the different eigenproblems satisfied the null constraint $\Proj^T \Sp \Proj = 0$ (we give the sum of the matrix elements, i.e, $\mathcal{A} = \sum_i \sum_j [\Proj^T \Sp \Proj]_{ij}$) and the corresponding criterion value ($\mathcal{B} = \trace(\Proj^T \Sn \Proj)$), when the orthogonalized projection vectors (i.e., the \ac{NCSDA} variant with Step~\ref{st:QR}) were used. For this experiment, we set the subspace dimensionality to 10. We show the results for the BU dataset as an example, while the results on other datasets behave in a similar manner.  

\begin{table*}[ht]
\caption{Average Precision for different \ac{NCSDA} variants}
\normalsize
\label{tab:nullresults}
\begin{tabular}{ll|cc|cc|cc|cc|cc}
 && \multicolumn{2}{c|}{\eqref{eq:nullSp}} & \multicolumn{2}{c|}{\eqref{eq:nullSn}} & \multicolumn{2}{c|}{\eqref{eq:nullSpSn}}  & \multicolumn{2}{c|}{\eqref{eq:nullSnSp}} & \multicolumn{2}{c}{\eqref{eq:nullSnSt}}\\
&& full D & 1-25D & full D & 1-25D & full D & 1-25D & full D & 1-25D & full D & 1-25D  \\
    \toprule
    \multirow{4}[2]{*}{JAFFE} & basic  & 0.792 & 0.722 & 0.465 & 0.436 & 0.712 & 0.370 & 0.792 & 0.741 & 0.710 & 0.576 \\
          & w/ St. \ref{st:spann} & 0.792 & \textbf{0.745} & 0.465 & 0.436 & 0.712 & 0.583 & 0.792 & 0.741 & 0.710 & 0.566 \\
          & w/ St.  \ref{st:QR} & 0.792 & 0.722 & 0.465 & 0.436 & 0.792 & 0.401 & 0.792 & 0.741 & 0.792 & 0.612 \\
          & w/ St. \ref{st:spann} + \ref{st:QR}  & 0.792 & \textbf{0.745} & 0.465 & 0.436 & 0.792 & 0.655 & 0.792 & 0.741 & 0.792 & 0.595 \\
    \midrule
    \multirow{4}[2]{*}{Kanade} & basic  & 0.361 & 0.343 & 0.357 & 0.347 & 0.263 & 0.244 & 0.362 & \textbf{0.500} & 0.245 & 0.280 \\
          & w/ St. \ref{st:spann} & 0.361 & 0.495 & 0.357 & 0.347 & 0.263 & 0.257 & 0.362 & \textbf{0.500} & 0.245 & 0.248 \\
          & w/ St.  \ref{st:QR} & 0.361 & 0.343 & 0.357 & 0.347 & 0.361 & 0.242 & 0.362 & \textbf{0.500} & 0.361 & 0.290 \\
          & w/ St. \ref{st:spann} + \ref{st:QR}  & 0.361 & 0.495 & 0.357 & 0.347 & 0.361 & 0.263 & 0.362 & \textbf{0.500} & 0.361 & 0.269 \\
    \midrule
    \multirow{4}[2]{*}{BU} & basic  & 0.379 & 0.399 & 0.313 & 0.323 & 0.231 & 0.201 & 0.379 & 0.480 & 0.231 & 0.208 \\
          & w/ St. \ref{st:spann} & 0.379 & \textbf{0.482} & 0.313 & 0.323 & 0.231 & 0.216 & 0.379 & 0.480 & 0.231 & 0.212 \\
          & w/ St.  \ref{st:QR} & 0.379 & 0.399 & 0.313 & 0.323 & 0.379 & 0.201 & 0.379 & 0.480 & 0.379 & 0.204 \\
          & w/ St. \ref{st:spann} + \ref{st:QR}  & 0.379 & \textbf{0.482} & 0.313 & 0.323 & 0.379 & 0.208 & 0.379 & 0.480 & 0.379 & 0.213 \\
    \midrule
    \multirow{4}[2]{*}{ORL} & basic  & 0.995 & 0.938 & 0.990 & 0.966 & 0.955 & 0.253 & 0.995 & \textbf{0.982} & 0.955 & 0.786 \\
          & w/ St. \ref{st:spann} & 0.995 & \textbf{0.982} & 0.990 & 0.966 & 0.955 & 0.801 & 0.995 & \textbf{0.982} & 0.955 & 0.782 \\
          & w/ St.  \ref{st:QR} & 0.995 & 0.938 & 0.990 & 0.966 & 0.995 & 0.273 & 0.995 & \textbf{0.982} & 0.995 & 0.804 \\
          & w/ St. \ref{st:spann} + \ref{st:QR}  & 0.995 & \textbf{0.982} & 0.990 & 0.966 & 0.995 & 0.801 & 0.995 & \textbf{0.982} & 0.995 & 0.796 \\
    \midrule
    \multirow{4}[2]{*}{AR} & basic  & 0.592 & 0.515 & 0.277 & 0.260 & 0.569 & 0.273 & 0.591 & 0.575 & 0.569 & 0.358 \\
          & w/ St. \ref{st:spann} & 0.592 & \textbf{0.577} & 0.277 & 0.260 & 0.569 & 0.372 & 0.591 & 0.575 & 0.569 & 0.371 \\
          & w/ St.  \ref{st:QR} & 0.592 & 0.515 & 0.277 & 0.260 & 0.592 & 0.270 & 0.592 & 0.575 & 0.592 & 0.376 \\
          & w/ St. \ref{st:spann} + \ref{st:QR}  & 0.592 & \textbf{0.577} & 0.277 & 0.260 & 0.592 & 0.383 & 0.592 & 0.575 & 0.592 & 0.375 \\
    \midrule
    \multirow{4}[2]{*}{Yale} & basic  & 0.729 & 0.730 & 0.150 & 0.144 & 0.445 & 0.176 & 0.729 & \textbf{0.731} & 0.413 & 0.309 \\
          & w/ St. \ref{st:spann} & 0.729 & 0.730 & 0.150 & 0.144 & 0.445 & 0.379 & 0.729 & \textbf{0.731} & 0.413 & 0.333 \\
          & w/ St.  \ref{st:QR} & 0.729 & 0.730 & 0.150 & 0.144 & 0.729 & 0.204 & 0.750 & \textbf{0.731} & 0.748 & 0.389 \\
          & w/ St. \ref{st:spann} + \ref{st:QR}  & 0.729 & 0.730 & 0.150 & 0.144 & 0.729 & 0.433 & 0.750 & \textbf{0.731} & 0.748 & 0.379 \\
    \midrule
		\multicolumn{1}{c}{\multirow{4}[2]{*}{15 scenes}} & basic  & 0.775 & 0.770 & 0.758 & 0.791 & 0.210 & 0.098 & 0.776 & \textbf{0.822} & 0.208 & 0.156 \\
          & w/ St. \ref{st:spann} & 0.775 & \textbf{0.822} & 0.758 & 0.791 & 0.210 & 0.171 & 0.776 & \textbf{0.822} & 0.208 & 0.163 \\
          & w/ St.  \ref{st:QR} & 0.775 & 0.770 & 0.758 & 0.791 & 0.775 & 0.099 & 0.776 & \textbf{0.822} & 0.775 & 0.173 \\
          & w/ St. \ref{st:spann} + \ref{st:QR}  & 0.775 & \textbf{0.822} & 0.758 & 0.791 & 0.775 & 0.219 & 0.776 & \textbf{0.822} & 0.775 & 0.174 \\
    \bottomrule


\multicolumn{12}{l}{\footnotesize{Eigenproblems compared for Step \ref{st:nullp} in Algorithm \ref{alg:NCSDA}:}}\\
\multicolumn{12}{l}{\footnotesize{\eqref{eq:nullSp}: $\Sp \projj = \lambda \projj \:\:\:\:$ \eqref{eq:nullSn}: $\Sn \projj = \lambda \projj \:\:\:\:$ \eqref{eq:nullSpSn}: $\Sp \projj = (\Sn + \mu\I)  \lambda \projj \:\:\:\:$ \eqref{eq:nullSnSp}: $\Sn \projj = (\Sp + \mu\I) \lambda \projj \:\:\:\:$ \eqref{eq:nullSnSt}: $\Sn \projj = \St \lambda \projj$}} 

\vspace{-7pt}
\end{tabular}
\end{table*}

\begin{table*}[ht]
\caption{Evaluation of constraint $\mathcal{A}$ and criterion $\mathcal{B}$ on BU dataset for \ac{NCSDA} variants using different eigenproblems for Step~\ref{st:nullp} followed by Step~\ref{st:QR} in Algorithm~\ref{alg:NCSDA}}
\normalsize
\label{tab:nullcrit}
\begin{tabular}{ll|cc|cc|cc|cc|cc}
&& \multicolumn{2}{c|}{\eqref{eq:nullSp}} & \multicolumn{2}{c|}{\eqref{eq:nullSn}} & \multicolumn{2}{c|}{\eqref{eq:nullSpSn}}  & \multicolumn{2}{c|}{\eqref{eq:nullSnSp}} & \multicolumn{2}{c}{\eqref{eq:nullSnSt}}    \\
&& full D & 10D & full D & 10D & full D & 10D & full D & 10D & full D & 10D \\

    \toprule
    \multicolumn{1}{l}{\multirow{2}[2]{*}{$\mathcal{A}$}} & Train & 1.6E-27 & 1.5E-29 & 335.74 & 26.75 & 2.7E-26 & 1.5E-28 & 4.6E-04 & 8.8E-07 & 3.1E-27 & 6.9E-31 \\
          & Test  & 125.71 & 0.41  & 202.04 & 14.27 & 124.54 & 0.02  & 89.37 & 1.16  & 127.97 & 0.04 \\
    \midrule
    \multicolumn{1}{r}{\multirow{2}[2]{*}{$\mathcal{B}$}} & Train & 60.68 & 5.60  & 173.56 & 110.35 & 60.68 & 0.17  & 60.72 & 18.28 & 60.68 & 0.44 \\
          & Test  & 17.48 & 2.05  & 65.26 & 46.34 & 17.48 & 0.05  & 17.49 & 6.89  & 17.48 & 0.10 \\
    \bottomrule
		

\multicolumn{10}{l}{\footnotesize{$\mathcal{A} = \sum_i \sum_j [\Proj^T \Sp \Proj]_{ij} \:\:\:\: \mathcal{B} = \trace(\Proj^T \Sn \Proj)$}} \\
\vspace{-10pt}
\end{tabular}
\end{table*}

We first see that the full dimensionality results of \eqref{eq:nullSn} ($\Sn \projj = \lambda \projj$) are clearly inferior to all the other variants. Table~\ref{tab:nullcrit} shows that, when \eqref{eq:nullSn} was used in Step~\ref{st:nullp}, the resulting projection vectors failed to satisfy the null constraint~$\mathcal{A}$. Furthermore, when we solved this eigenproblem to obtain vectors spanning the null space of $\tildeSn$, we observed that these vectors are not truly orthogonal to the vectors spanning the null space of $\tildeSp$ solved from \eqref{eq:nullSp} (i.e., $\Nullpm^T\Nullnm \neq 0$, where $\Nullpm$ and $\Nullnm$ are matrices containing as their columns the vectors spanning $\Nullp$ and $\Nulln$). As a result, the row space of $\tildeSn$ solved from \eqref{eq:nullSn} is not accurately aligned with the null space of $\tildeSp$ and the ability of the solution to satisfy the constraint of $\mathcal{J}_N$ suffers making the classification performance worse as well. We will come back to this discrepancy in the next Section~\ref{sssec:orthogexperiments} and now concentrate on other variants of \ac{NCSDA}. 

When we took full \ac{NCSDA} dimensionality as our final subspace dimensionality and orthogonalized the projection matrices (i.e., the full D w/ St. \ref{st:QR}), the results for the other eigenproblems for Step~\ref{st:nullp} were almost equal. Table~\ref{tab:nullcrit} shows that all these approaches manage to approximately satisfy the constraint~$\mathcal{A}$ and result in similar values for the criterion $\mathcal{B}$ confirming that the projection matrices span approximately the same subspace. Furthermore, we see from Table~\ref{tab:nullresults} that orthogonality of the projection vectors is indeed beneficial. The results for the originally unorthogonal projection vectors resulting from \eqref{eq:nullSpSn} and \eqref{eq:nullSnSt} (i.e., the full D Basic) were clearly improved by orthogonalization.

Since the innate \ac{NCSDA} dimensionality is typically too high, we are more interested in the results with dimensionality reduced to 1-25. Therefore, we have highlighted in Table~\ref{tab:nullresults} the best result with a reduced dimensionality for each dataset. The results for the Basic variant confirm our theoretical evaluation on the ability of \eqref{eq:nullSnSp} to rank the projection vectors for dimensionality reduction better than the other eigenproblems. As expected, the low-dimensional results obtained with \eqref{eq:nullSpSn} and \eqref{eq:nullSnSt} are much worse. Interestingly, the eigenproblem \eqref{eq:nullSp} showed a better ability to rank vectors than \eqref{eq:nullSpSn} or \eqref{eq:nullSnSt}. While only eigenvectors corresponding to zero eigenvalues are selected for \eqref{eq:nullSp}, in practice, all the eigenvalues are not exactly zero but values of order $10^{-16}$. Clearly, ranking these values also provides some useful information for ranking while not as much as \eqref{eq:nullSnSp}.

In accordance with \eqref{eq:trace}, Step~\ref{st:spann} had no effect on the full dimensionality performance. As expected, it could improve the low-dimensional results of \eqref{eq:nullSp}. It could not help with \eqref{eq:nullSpSn} and \eqref{eq:nullSnSt} due to the eigenvalue equalization resulting from the $\tildeSn$/$\tildeSt$-orthogonality. The projection vectors resulting from \eqref{eq:nullSn} and \eqref{eq:nullSnSp} were already ranked according to their ability to maximize $\mathcal{B}$ and Step~\ref{st:spann} had no effect. Step~\ref{st:QR} did not affect the already orthogonal projection vectors resulting from \eqref{eq:nullSp}, \eqref{eq:nullSn}, and \eqref{eq:nullSnSp}, but it improved the results when applied with \eqref{eq:nullSpSn} and \eqref{eq:nullSnSt}.      

Table~\ref{tab:nullcrit} shows that \eqref{eq:nullSpSn} and \eqref{eq:nullSnSt} satisfied the null constraint $\mathcal{A}$ well, but they failed to rank projection vectors for dimensionality reduction leading to very small values of criterion $\mathcal{B}$. The eigenproblem \eqref{eq:nullSnSp} maximized $\mathcal{B}$ much better, while \eqref{eq:nullSn} produced even higher values for $\mathcal{B}$ but with the cost of failing to satisfy the null constraint $\mathcal{A}$. These results also generalized well for the unseen test samples.

As a conclusion, we select as our representative \ac{NCSDA} method for the following comparative experiments to be the approach where \eqref{eq:nullSnSp} is used in Step~\ref{st:nullp}, while the optional Steps~\ref{st:spann} and \ref{st:QR} are avoided. This approach is consistently among the top performing variants.  

Here, we want to point out that for our analysis of \ac{NCSDA} it is important to exploit the symmetry of the scatter matrices in the implementation of eigenanalysis of Step \ref{st:nullp}. 
If $\tildeSp$ or $\tildeSn$ are not exactly symmetric for \eqref{eq:nullSp} or \eqref{eq:nullSn} (i.e, they have numerical errors of order $10^{-20}$), Matlab's \texttt{eig}-function does not produce real eigenvalues and orthogonal eigenvectors. This, in turn, makes the classification performance dramatically worse. We observe that while Matlab ensures that $\texttt{A*A'}$ is exactly symmetric, this is not the case for expressions like $\texttt{A(:, inds)*A(:, inds)'}$. Therefore, it is important to create separate matrices for the positive and negative samples and not just index the total data matrix when forming the corresponding scatter matrices. 
For \eqref{eq:nullSpSn}-\eqref{eq:nullSnSt}, an eigenanalysis algorithm preserving the symmetry properties should be used. While QZ-algorithm \cite{Moler1973QZ} is numerically more stable, it does not exploit the symmetric structure of the matrices, which typically results in complex eigenvalues and vectors. Cholesky method \cite{Davies2001Cholesky} is specifically meant for symmetric matrices and ensures the properties discussed above. 

\subsubsection{Experiments on Uncorrelated, Orthogonal, and Regularized Orthogonal \ac{CSDA}} \label{sssec:orthogexperiments}

We now turn our attention to \ac{UCSDA}, \ac{OCSDA}, and \ac{ROCSDA} discussed in Section~\ref{ssec:OCSDA}. In Table~\ref{tab:orthresults}, we give Average Precisions for \ac{UCSDA}, \ac{OCSDA}, and \ac{ROCSDA} with different ways to solve Step~\ref{st:spannO} in Algorithm~\ref{alg:OCSDA}. For comparison, we also applied an algorithm similar to \ac{UCSDA} but not involving whitening (w/o wh.). Note, that the resulting algorithm gives orthogonal projection vectors without needing the orthogonalizing of Step~\ref{st:QRO}. Furthermore, we carried out an evaluation on the null constraint $\mathcal{A}$ and criterion $\mathcal{B}$ as in Table~\ref{tab:nullcrit} for \ac{OCSDA} and \ac{ROCSDA}. These results are given in Table~\ref{tab:orthcrit}.

The version with no whitening is equivalent to Basic \ac{NCSDA}. The variant using \eqref{eq:nullSnSp} is exactly \ac{NCSDA} using \eqref{eq:nullSnSp}, while the variants using \eqref{eq:svdn} and \eqref{eq:svdp} are almost identical to \ac{NCSDA} using \eqref{eq:nullSn} and \eqref{eq:nullSp}, respectively. The only difference is that the former use \ac{SVD} to solve the eigenproblems in \eqref{eq:nullSn} and \eqref{eq:nullSp}. We see that the full dimensional results are equal and, for \eqref{eq:svdn}/\eqref{eq:nullSn}, the results with dimensionality reduction are also the same. For \eqref{eq:svdp}/\eqref{eq:nullSp}, the results with reduced dimensionality have some variation, which is reasonable as the projection vectors correspond to zero eigenvalues having no particular ranking. 

For the version without whitening (as for \ac{NCSDA}), the results based purely on the eigenvectors of $\tildeSn$ are clearly worse than the results obtained with the other two variants. With \ac{UCSDA}, the results for all the variants are equalized but also clearly deteriorated. This can be partially explained by the unorthogonal projection vectors, as the results with full dimensionality are clearly better for \ac{OCSDA}, where the only difference to  \ac{UCSDA} is the orthogonalization applied at the end. However, also for \ac{OCSDA} the results after dimensionality reduction are poor. As explained in Section~\ref{ssec:OCSDA}, $\St$-orthogonality following from the whitening makes the eigenvalues useless for ranking the vectors. This can be cured by regularization of the \ac{PCA} whitening  introduced in \ac{ROCSDA}. With the tiny value of $10^{-7}$ added to the eigenvalues of $\St$ in the whitening, the difference in the results with the dimensionality reduction is dramatic. In fact, it turns out that the projection vector ranking of \ac{ROCSDA} outperforms all the approaches considered with \ac{NCSDA}. Now the results are also equal for \eqref{eq:svdn} and \eqref{eq:nullSnSp} as they theoretically should be.  We believe this is achieved by the whitening step which can prevent the numerical errors caused by the previously ill-conditioned $\St$ matrix. \ac{ROCSDA} did not help with \eqref{eq:svdp} using the eigenvectors corresponding to zero eigenvalues, which cannot be ranked in any case.

Table~\ref{tab:orthcrit} supports our conclusions: With whitening applied, the eigenvectors spanning the row space of $\tildeSn$ obtained from \eqref{eq:svdn} can also satisfy the null constraint $\mathcal{A}$. \ac{OCSDA} cannot rank the projection vectors, which is demonstrated by the low values of criterion $\mathcal{B}$ for the solution with reduced dimensionality. \ac{ROCSDA} can significantly improve the criterion value, when \eqref{eq:svdn} or \eqref{eq:nullSnSp} is used in Step~\ref{st:spannO}.
  
\subsubsection{Comparative experiments}

At last, we compared the proposed methods against standard \ac{CSDA} as well as the most recent variants. Table~\ref{tab:comparisons} shows the results of \ac{CSDA} \cite{goudelis2007classspecific}, \ac{CSSR} \cite{arashloo2014csksr,iosifidis2015CSRDA}, \ac{CSSR} using Cholesky decomposition \cite{iosifidis2017cskdaRev}, Probabilistic \ac{CSDA} \cite{iosifidis2018probabilistic} along with the results for the proposed methods \ac{NCSDA} (from Table~\ref{tab:nullresults}), \ac{ROCSDA} (from Table~\ref{tab:orthresults}), \ac{HNCSDA}, and \ac{HOCSDA}. 

Here, it should be noted that we have used for \ac{CSDA} a standard implementation using QZ-method for eigenanalysis. For a full rank $\St$ matrix (as is the case with \ac{NPT}), \ac{NCSDA} using \ref{eq:nullSnSp} in Step~\ref{st:nullp} and further dimensionality reduction actually reduces to the standard \ac{CSDA}. If we make sure that the scatter matrices are exactly symmetric and use Cholesky method for eigenanalysis, there is no difference between \ac{CSDA} and the selected \ac{NCSDA} version. \ac{ROCSDA}, however, can further improve these results on most datasets.

The recent \ac{CSDA} variants exploiting Spectral Regression and the probabilistic formulation clearly outperform the standard \ac{CSDA} and its straightforward extensions. Nevertheless, the when the concepts of null space analysis are combined with the heterogeneous formulation for the negative class as in \ac{HNCSDA} and \ac{HOCSDA}, the results are comparable to or even outperform these recent \ac{CSDA} variants. 

\section{Conclusions}\label{sec:conclusions}
In this paper, we present a null space analysis for \acl{CSDA} and present a number of extensions exploiting the concepts. We provide a theoretical and experimental analysis on the significance of each algorithmic step and show that a proper exploitation of the null space concepts can lead to improved results on the baseline \ac{CSDA}. Furthermore, we propose to combine the null space concepts with a recently proposed multi-modal formulation for the negative class scatter and show that this approach can produce results that are comparable to or even outperform the most recent \ac{CSDA} variants. 


\begin{table}[htpb]
  \centering
  \caption{Average Precision for different \ac{UCSDA}, \ac{OCSDA}, and \ac{ROCSDA} variants}
	\normalsize
	\addtolength{\tabcolsep}{-4pt}  
    \begin{tabular}{ll|cc|cc|cc}
		&& \multicolumn{2}{c|}{\eqref{eq:svdn}} & \multicolumn{2}{c|}{\eqref{eq:svdp}} & \multicolumn{2}{c}{\eqref{eq:nullSnSp}}  \\
		&& full D & 1-25D & full D & 1-25D & full D & 1-25D \\
\toprule
    \multirow{4}[2]{*}{JAFFE} & w/o wh. & 0.465 & 0.436 & 0.792 & 0.704 & 0.792 & 0.741 \\
          & \ac{UCSDA} & 0.710 & 0.579 & 0.710 & 0.645 & 0.710 & 0.621 \\
          & \ac{OCSDA} & 0.792 & 0.607 & 0.792 & 0.662 & 0.792 & 0.649 \\
          & \ac{ROCSDA} & 0.792 & \textbf{0.755} & 0.792 & 0.664 & 0.792 & \textbf{0.755} \\
    \midrule
    \multirow{4}[2]{*}{Kanade} & w/o wh. & 0.357 & 0.347 & 0.361 & 0.341 & 0.362 & 0.500 \\
          & \ac{UCSDA} & 0.245 & 0.229 & 0.245 & 0.271 & 0.245 & 0.237 \\
          & \ac{OCSDA} & 0.361 & 0.248 & 0.361 & 0.272 & 0.361 & 0.225 \\
          & \ac{ROCSDA} & 0.361 & \textbf{0.510} & 0.361 & 0.265 & 0.361 & \textbf{0.510} \\
    \midrule
    \multirow{4}[2]{*}{BU} & w/o wh. & 0.313 & 0.323 & 0.379 & 0.336 & 0.379 & 0.480 \\
          & \ac{UCSDA} & 0.231 & 0.215 & 0.231 & 0.229 & 0.231 & 0.201 \\
          & \ac{OCSDA} & 0.379 & 0.214 & 0.379 & 0.231 & 0.379 & 0.202 \\
          & \ac{ROCSDA} & 0.379 & \textbf{0.482} & 0.379 & 0.226 & 0.379 & \textbf{0.482} \\
    \midrule
    \multirow{4}[2]{*}{ORL} & w/o wh. & 0.990 & 0.966 & 0.995 & 0.924 & 0.995 & \textbf{0.982} \\
          & \ac{UCSDA} & 0.955 & 0.796 & 0.955 & 0.908 & 0.955 & 0.692 \\
          & \ac{OCSDA} & 0.995 & 0.807 & 0.995 & 0.912 & 0.995 & 0.722 \\
          & \ac{ROCSDA} & 0.995 & \textbf{0.982} & 0.995 & 0.917 & 0.995 & \textbf{0.982} \\
    \midrule
    \multirow{4}[2]{*}{AR} & w/o wh. & 0.277 & 0.260 & 0.592 & 0.530 & 0.591 & \textbf{0.575} \\
          & \ac{UCSDA} & 0.569 & 0.358 & 0.569 & 0.470 & 0.569 & 0.367 \\
          & \ac{OCSDA} & 0.592 & 0.340 & 0.592 & 0.484 & 0.592 & 0.380 \\
          & \ac{ROCSDA} & 0.592 & 0.563 & 0.592 & 0.480 & 0.592 & 0.563 \\
    \midrule
    \multirow{4}[2]{*}{Yale} & w/o wh. & 0.150 & 0.144 & 0.729 & 0.726 & 0.729 & 0.731 \\
          & \ac{UCSDA} & 0.413 & 0.207 & 0.418 & 0.666 & 0.418 & 0.385 \\
          & \ac{OCSDA} & 0.748 & 0.320 & 0.729 & 0.660 & 0.748 & 0.434 \\
          & \ac{ROCSDA} & 0.748 & \textbf{0.783} & 0.729 & 0.662 & 0.748 & \textbf{0.783} \\
    \midrule
		\multirow{4}[2]{*}{15 sc.} & w/o wh. & 0.758 & 0.791 & 0.775 & 0.704 & 0.776 & 0.822 \\
          & \ac{UCSDA} & 0.208 & 0.140 & 0.208 & 0.332 & 0.208 & 0.099 \\
          & \ac{OCSDA} & 0.775 & 0.161 & 0.775 & 0.341 & 0.775 & 0.109 \\
          & \ac{ROCSDA} & 0.775 & \textbf{0.829} & 0.775 & 0.340 & 0.775 & \textbf{0.829} \\
    \bottomrule
		

\multicolumn{8}{l}{\footnotesize{Approaches compared for Step \ref{st:spann} in Algorithm \ref{alg:OCSDA}:}} \\
\multicolumn{8}{l}{\footnotesize{\eqref{eq:svdn}: \ac{SVD} on $\Data_n \:\:\:\:$ \eqref{eq:svdp}: \ac{SVD} on $\Data_p$ \eqref{eq:nullSnSp}: $\Sn \projj = (\Sp + \mu\I) \lambda \projj$ }}
\vspace{-7pt}
    \end{tabular}%
		\addtolength{\tabcolsep}{4pt}  
  \label{tab:orthresults}%
\end{table}%

\begin{table*}[htbp]
  \centering
  \caption{Evaluation of constraint $\mathcal{A}$ and criterion $\mathcal{B}$ on BU dataset for \ac{OCSDA} and \ac{ROCSDA} variants using different approaches for Step~\ref{st:spannO} in Algorithm~\ref{alg:OCSDA}}
	\normalsize
	\addtolength{\tabcolsep}{-2pt}  
    \begin{tabular}{ll|cc|cc|cc|cc|cc|cc}
		&& \multicolumn{6}{c|}{$\mathcal{A} = \sum_i \sum_j [\Proj^T \Sp \Proj]_{ij}$} & \multicolumn{6}{c}{$\mathcal{B} = \trace(\Proj^T \Sn \Proj)$} \\
		&& \multicolumn{2}{c|}{\eqref{eq:svdn}} & \multicolumn{2}{c|}{\eqref{eq:svdp}} & \multicolumn{2}{c|}{\eqref{eq:nullSnSp}} & \multicolumn{2}{c|}{\eqref{eq:svdn}} & \multicolumn{2}{c|}{\eqref{eq:svdp}} & \multicolumn{2}{c}{\eqref{eq:nullSnSp}}\\  
		&& full D & 10D & full D & 10D & full D & 10D & full D & 10D & full D & 10D & full D & 10D \\

    \toprule
    \multirow{2}[2]{*}{Train} & \ac{OCSDA} & 1.1E-26 & 6.6E-30 & 3.1E-27 & 1.2E-30 & 2.7E-27 & 8.0E-31 & 60.68 & 0.38  & 60.68 & 1.68  & 60.68 & 0.44 \\
          & \ac{ROCSDA} & 1.9E-11 & 3.1E-13 & 3.2E-27 & 2.5E-30 & 1.9E-19 & 3.1E-21 & 60.68 & 15.34 & 60.68 & 1.68  & 60.68 & 15.34 \\
    \midrule
    \multirow{2}[2]{*}{Test} & \ac{OCSDA} & 123.63 & 0.03  & 125.96 & 0.15  & 128.07 & 0.04  & 17.48 & 0.09  & 17.48 & 0.42  & 17.48 & 0.10 \\
          & \ac{ROCSDA} & 93.92 & 1.00  & 125.96 & 0.15  & 93.92 & 1.00  & 17.48 & 5.87  & 17.48 & 0.42  & 17.48 & 5.87 \\
    \bottomrule
		\vspace{-7pt}
    \end{tabular}%
		\addtolength{\tabcolsep}{2pt}  
  \label{tab:orthcrit}%
\end{table*}%

\begin{table*}[ht]
\caption{Average Precision for the proposed methods and competing \ac{CSDA} variants}
\normalsize
\label{tab:comparisons}
\begin{tabular}{l|cccccccc}

 & \ac{CSDA} & \bf{\ac{NCSDA}} & \bf{\ac{ROCSDA}} & \ac{CSSR}  & Chol \ac{CSSR} & Prob \ac{CSDA} & \bf{\ac{HNCSDA}} & \bf{\ac{HOCSDA}}  \\
\hline
    JAFFE & 0.729 & 0.741 & 0.755 & 0.888 & 0.888 & 0.868 & \textbf{0.889} & 0.887 \\
    Kanade & 0.495 & 0.500 & 0.510 & 0.674 & 0.660 & 0.662 & \textbf{0.680} & 0.671 \\
    BU    & 0.480 & 0.480 & 0.482 & 0.628 & \textbf{0.630} & 0.616 & 0.615 & 0.625 \\
    ORL   & 0.982 & 0.982 & 0.982 & \textbf{0.999} & \textbf{0.999} & 0.998 & \textbf{0.999} & 0.998 \\
    AR    & 0.565 & 0.575 & 0.563 & 0.985 & 0.981 & 0.976 & 0.984 & \textbf{0.988} \\
    Yale  & 0.719 & 0.731 & 0.783 & 0.912 & 0.885 & \textbf{0.955} & 0.914 & 0.884 \\
    15 scenes & 0.818 & 0.822 & 0.829 & 0.859 & \textbf{0.860} & 0.859 & \textbf{0.860} & \textbf{0.860} \\
\vspace{-7pt}
\end{tabular}
\end{table*}

\bibliographystyle{myIEEEtran}
\bibliography{bibliography}

\end{document}